%% file: main.tex
\documentclass[11pt]{article}

\usepackage[preprint]{acl}

\usepackage{times}
\usepackage{latexsym}

\usepackage[T1]{fontenc}
\usepackage[utf8]{inputenc}

\usepackage{microtype}

\usepackage{inconsolata}

\usepackage{graphicx}

\usepackage{subcaption}
\usepackage{booktabs}
\usepackage{amssymb}
\usepackage{mdframed}
\usepackage{xcolor}
\usepackage{tikz}
\usetikzlibrary{arrows.meta, positioning, shapes.geometric, calc, fit, backgrounds, decorations.pathreplacing}

\newenvironment{prompt}
  {\begin{mdframed}[
    backgroundcolor=gray!10,
    linecolor=gray!30,
    linewidth=0.5pt,
    innertopmargin=8pt,
    innerbottommargin=8pt,
    leftmargin=0pt,
    rightmargin=0pt,
    skipabove=\baselineskip,
    skipbelow=\baselineskip
  ]\fontsize{8.5}{10.4}\selectfont\ttfamily}
  {\end{mdframed}}

\title{Disentangling and Mitigating Biases in LLM-as-a-Judge Evaluation of Summaries\thanks{Accepted for publication at the EACL 2027 main conference.}}
\author{
  \textbf{Alireza Zarei\textsuperscript{1,2}},
  \textbf{Jonathn Chang\textsuperscript{3}},
  \textbf{Radoslav Koynov\textsuperscript{1,2}}, \\
  \textbf{Rosan Sharma\textsuperscript{4}},
  \textbf{Andreas Henrich\textsuperscript{5}},
  \textbf{Philipp Wieder\textsuperscript{1,2}} \\
  \textsuperscript{1}GWDG \quad
  \textsuperscript{2}University of Göttingen \quad
  \textsuperscript{3}Cornell University \\
  \textsuperscript{4}Freie Universität Berlin \quad
  \textsuperscript{5}University of Bamberg \\
  \small{\textbf{Correspondence:} \href{mailto:azarei@gwdg.de}{azarei@gwdg.de}}
}

\begin{document}
\maketitle

\begin{abstract}
Large language models (LLMs) increasingly both generate and evaluate summaries, but fair comparison is difficult because judge bias can dominate genuine quality differences. We summarize 200 recent Guardian Science articles with four open-weight LLMs and compare them with four LLM judges. Length dominates uncontrolled comparisons, with the longer summary winning 74\% of decided pairs, a share that drops to 57\% once a fixed length budget is enforced with Divide--Verify--Refine. Position then takes over -- under forced choice, order-swapped comparisons flip on up to 53\% of pairs, and the favored slot is the one adjacent to the article, surfacing as recency or primacy according to prompt layout -- yet the pairs a judge flips on are only negligibly more similar than those it ranks consistently. This is consistent with position serving as a fallback when quality differences are difficult to distinguish. Combining an Equal option with reflection reduces total order disagreement by 44--65\% in all eight judge--placement settings. These outcomes are label-free diagnostics of judge invariance. They measure self-consistency rather than agreement with human preference. Within this setting, they argue for designing LLM summarization evaluation jointly around candidate-side length control and order-swapped judge protocols combining an Equal option with reflection.
\end{abstract}

\section{Introduction}

Instruction-tuned large language models (LLMs) have made abstractive summarization a natural application, since they generate fluent summaries from simple instructions, adapt to different domains, and support user-facing controls such as length, tone, and level of detail. Recent surveys and benchmarks report that LLM-based summarizers can outperform previous abstractive systems~\cite{shakil2024abstractive,zhang2024benchmarking} and that their summaries are often preferred to reference summaries on news datasets~\cite{zhang2024benchmarking,pu2023summarization}. At the same time, controllable summarization remains imperfect even for strong instruction-tuned models~\cite{liu2024benchmarking}, and output length alone can strongly affect perceived quality in preference judgments~\cite{singhal2024longwaygoinvestigating,saito2023verbositybiaspreferencelabeling}.

These developments shift the central difficulty from generating fluent summaries to evaluating them reliably. Reference-based metrics such as ROUGE~\cite{lin2004rouge} and BERTScore~\cite{zhang2019bertscore} require gold summaries and score similarity to a reference, whether by surface overlap or by embedding similarity. Human evaluation does not require a reference summary but is expensive and sensitive to protocol design~\cite{liu2023revisiting} and to annotator variation~\cite{harman2004effects}. LLM-as-a-judge evaluation offers a scalable alternative, with prior work showing strong alignment with human judgments in some settings~\cite{liu2023g,zheng2023judging}. However, LLM judges themselves exhibit systematic biases and instability when comparing high-quality outputs~\cite{wang2024large,saito2023verbositybiaspreferencelabeling,panickssery2024llm,shen2023large}.

\begin{figure*}[!t]
\centering
\resizebox{\linewidth}{!}{\input{figures/framework}}
\caption{Study design and main results. \textbf{(a)}~Summaries are length-controlled
with Divide--Verify--Refine~\cite{zhang2025divide} and judged in both orders under
article-first and summaries-first layouts; red marks the article-adjacent slot.
\textbf{(b)}~Swap-pair outcomes: A/B denote presentation slots and E denotes Equal.
Selecting the same model twice is an order-independent preference. Total order
disagreement comprises selecting the first slot twice (primacy) or the second
slot twice (recency), which produces strict preference reversals, and Equal in
exactly one order (tie/preference disagreement). Equal in
both orders is stable abstention, excluded from disagreement.
\textbf{(c)}~Factorial ablation of Equal and reflection within per-criterion
judging. Values show total order disagreement ranges across eight judge--layout
settings; arrows show within-setting changes in percentage points (negative means
lower disagreement). \textbf{(d)}~Disagreement by criterion with both mitigations
enabled; bars show medians across the same eight settings. Table~\ref{tab:winner_coverage}
provides the component breakdown.}
\label{fig:framework}
\end{figure*}
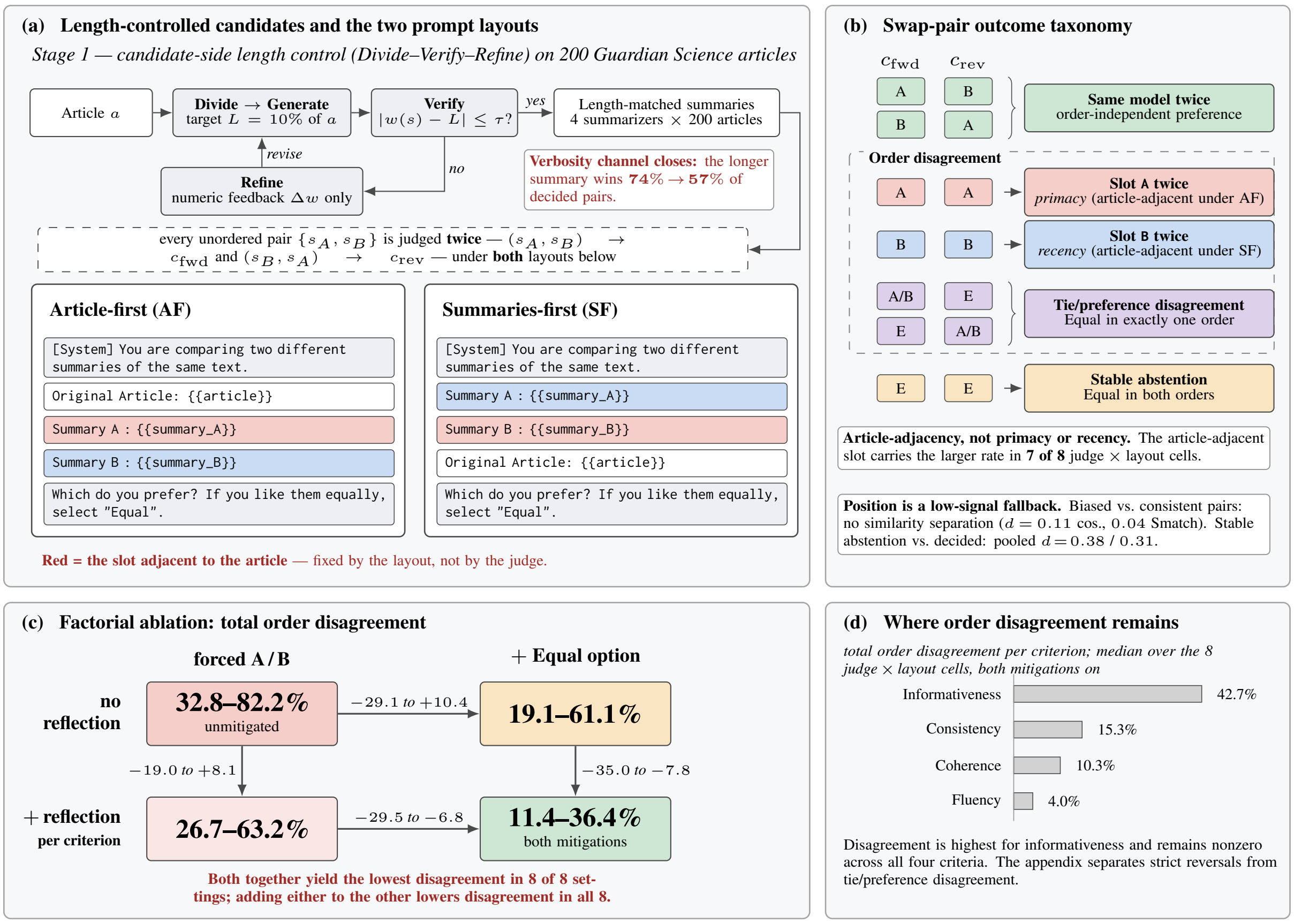

In this paper, we study LLM judges through invariance diagnostics of their news summarization judgments. Reversing the presentation order of two summaries, or moving the article, cannot change which summary is better, so a verdict that changes with either is wrong in at least one presentation, whatever the true ranking is. These diagnostics need no gold labels, and they do not establish which summarizer is better or whether the judgments agree with human preference. We construct a dataset of 200 recent Guardian Science articles, summarize them with four open-weight LLMs, and ask which factors actually drive a judge's verdict. Length comes first. The longer summary wins most decided comparisons, so we constrain every summary to $10\%$ of its article with Divide--Verify--Refine (DVR)~\cite{zhang2025divide}, which removes most of the verbosity channel. Once length is equalized, position takes over. We measure it with order-swapped comparisons -- the same two summaries presented in both orders -- and separate primacy from recency under two prompt layouts, varying whether the article precedes or follows the summaries. The direction tracks the layout rather than the judge, as the favored slot is the one adjacent to the article. To mitigate position bias, we combine an explicit \emph{Equal} option with reflection-based per-criterion judging and assess their effects through a factorial ablation. Figure~\ref{fig:framework} summarizes the design and the main results. We make the following contributions.
\begin{enumerate}
    \item A recent, post-cutoff news summarization dataset of 200 Guardian Science articles with summaries from four open-weight LLMs.
    \item A quantification of verbosity bias on uncontrolled drafts and the application of a Divide--Verify--Refine length budget~\cite{zhang2025divide} that removes most of the verbosity channel from judge preferences.
    \item A characterization of order sensitivity in length-controlled summary judging, distinguishing changes in the preferred summary from changes between preference and abstention, and showing how prompt layout shapes directional position bias.
    \item A factorial ablation of two mitigations -- an Equal option and reflection-based per-criterion judging -- showing that combining an Equal option with reflection yields the lowest observed total order disagreement in all eight judge--placement settings.
\end{enumerate}

  % \item We supplement aggregate preference scores with knowledge-type and AMR-based diagnostics to analyze which facts and semantic structures different summarizers preserve.

\section{Related work}

LLM evaluators are widely used for scalable NLG assessment. G-Eval~\cite{liu2023g} shows strong alignment with human judgments, and MT-Bench/Chatbot Arena popularize pairwise LLM judging~\cite{zheng2023judging}. However, prior work shows that LLM judges can exhibit position bias~\cite{wang2024large}, verbosity bias~\cite{saito2023verbositybiaspreferencelabeling}, self-preference bias~\cite{panickssery2024llm}, and instability on high-quality summarization systems~\cite{shen2023large}. Recent surveys catalog these biases across tasks~\cite{li2024llmsasjudges}, and \citet{shi2025judging} study position bias systematically across judges. We directly measure position, verbosity, and self-preference bias in pairwise summary judgments. On expert-knowledge tasks, LLM judges agree with domain experts only about two thirds of the time~\cite{szymanski2025limitations}, which motivates our interpretation of the rates as judge-consistency diagnostics rather than accuracy measures. Human-evaluation studies also show that summarization judgments depend on the evaluation protocol~\cite{liu2023revisiting} and exhibit large annotator variation~\cite{harman2004effects}. We therefore measure order-swap flips directly and compare bias-mitigation strategies including an Equal option and reflection-based judging. Aggregate preference scores can obscure why two summaries differ. Broad automatic metrics have known limitations~\cite{fabbri2021summeval}, and even dedicated factual-consistency metrics remain imperfect~\cite{laban2022summac,fabbri2022qafacteval}. We therefore also characterize each evaluated pair using structural similarity measured by AMR Smatch~\citep{cai-knight-2013-smatch}.

\section{Experimental setup}
\label{sec:setup}

To minimize the risk of data contamination \cite{li2024task}, we focus on LLM summarization for recent, unevaluated data. The Guardian provides programmatic API access to recent articles. We selected its Science pillar because the density of facts and named entities makes content selection consequential under a tight length budget. This is a methodological rationale for the test bed, not evidence that science is the hardest news domain for judges. Whether the observed bias magnitudes transfer to other sources or domains remains open. We therefore created a new dataset of 200 articles from The Guardian's Science pillar published between 6 July 2025 and 29 October 2025.

To establish a diverse cohort for evaluation, we select four large open-weight LLMs from four different providers, incorporating models with varied multilingual and regional alignments. GPT-OSS-120b is intentionally included both as a summary generator and as an evaluator, allowing us to evaluate the impact of self-preference~\cite{panickssery2024llm}. We observe no aggregate self-preference relative to the other judges in this setting (Appendix~\ref{app:self_preference}). All models use greedy decoding settings (temperature~$0$, top-$p=1$) and a maximum output budget of 32{,}768 tokens, where supported. These settings do not make the two reasoning judges deterministic, and Appendix~\ref{app:repeats} measures how much their verdicts move between identical calls. We use chain-of-thought (CoT) prompting~\cite{wei2022chain} to explicitly allow the model to take a series of intermediate reasoning steps before generating the final text (Appendices~\ref{app:basic_prompt} and~\ref{app:pairwise_prompt}). CoT is a standard baseline here rather than an optimized choice, since prompting is an instrument of this study, not its object, and alternatives such as Tree-of-Thoughts~\cite{yao2023tree}, Program-of-Thoughts~\cite{chen2023program} and Plan-and-Solve~\cite{wang2023plan} target multi-step search and arithmetic rather than a pairwise verdict. We instead vary the judging protocol itself (Section~\ref{sec:position_mitigation}) and bound the effect of prompt wording on the judge separately (Appendix~\ref{app:prompt_sensitivity}).

\subsection{Summarizer models} We generate candidate summaries with four open-weight models, GPT-OSS-120b (OpenAI, 120B parameters; \citealp{openai2025gptoss}), Kimi-K2-Instruct-0905 (MoonshotAI, 1040B parameters; \citealp{kimiteam2025kimik2}), GLM-4.5-Air-FP8 (Z.ai, 106B parameters; \citealp{glmteam2025glm45}), and Qwen3-235B-A22B-Instruct-2507 (Alibaba, 235B parameters, 22B active; \citealp{qwenteam2025qwen3}). Kimi-K2-Instruct-0905 and Qwen3-235B-A22B-Instruct-2507 are non-reasoning instruction models, GLM-4.5-Air is a hybrid thinking model, and GPT-OSS-120b performs reasoning with medium effort. In all cases, we retain only the summary.

\subsection{LLM judges} We evaluate the summaries using four judges, GPT-4o-mini (\citealp{openai2024gpt4omini}), GPT-5-mini (OpenAI, reasoning\_effort=minimal, verbosity=low; \citealp{openai2025gpt5developers}), GPT-OSS-120b (OpenAI, 120B parameters, reasoning\_effort=medium -- the inference provider's default, which Appendix~\ref{app:repeats} verifies; \citealp{openai2025gptoss}) and DeepSeek-V3.1 (DeepSeek, 671B parameters, FP8; \citealp{deepseekai2024deepseekv3,deepseekai2025deepseekv31}). The inference budget of the full grid is reported in Appendix~\ref{app:compute} and package versions are listed in Appendix~\ref{app:versions}.

\subsection{Pairwise preference protocol}

To evaluate the summaries based on their overall quality, we elicit pairwise preferences between every pair of summarizer models on the set of 200~articles. For every article and every ordered pair $(A,B)$ of summarizer models, the judge is shown the article together with the two candidate summaries (labeled \emph{A} and \emph{B}) and is asked to (i)~return a verdict in $\{A, B\}$ for overall quality, and (ii)~assign a preference-strength score $d \in \{1,2,3\}$ ($1$ = slight, $2$ = clear, $3$ = decisive). The same prompt is used for all judges (Appendix~\ref{app:pairwise_prompt}). With 4~summarizer models, 6~unordered model pairs, 200~articles, and both orderings, this yields $2{,}400$ pairwise judgments per judge. 

\section{Verbosity bias}
\label{sec:verbosity_bias}

Previous research has shown that human evaluators perceive longer summaries as higher quality~\cite{stiennon2020learning} and that LLM judges prefer longer answers~\cite{saito2023verbositybiaspreferencelabeling}. 
We confirm a strong association between summary length and judge preference on our data. When the four judges compare the models' initial (uncontrolled-length) drafts, the longer summary wins $\approx\!74\%$ of decided pairs on average, with a mean winner--loser gap of $\sim$13--14~words and a point-biserial correlation between the length gap and the (dichotomous) verdict of $r_{pb}\approx+0.25$ ($p<10^{-19}$ for every judge).

\subsection{Verbosity bias mitigation}
\label{sec:verbosity_mitigation}

Asking for a specific word count does not ensure compliance. Across the 200 articles, first drafts land inside the $\pm10\%$ target band only $60\%$ of the time for GPT-OSS-120b, $50\%$ for Qwen3-235B, $20\%$ for Kimi-K2 and $7\%$ for GLM-4.5-Air (Appendix~\ref{app:dvr_convergence}). We therefore enforce the length budget with the \textbf{Divide--Verify--Refine (DVR)} framework of \citet{zhang2025divide}, shown in panel~(a) of Figure~\ref{fig:framework}. The instruction is \emph{divided} into a single verifiable word-count constraint, a deterministic word counter \emph{verifies} each draft, and on violation the model is asked to \emph{refine} it, using only numeric feedback on the draft's word count, the target, and the deviation $\Delta w$. The exact refinement message and the loop's restart schedule are given in Appendix~\ref{app:dvr_prompt}. The loop repeats until the constraint holds, subject to a safety cap of $100$ generations per article, after which the closest draft produced so far is kept. The cap is reached for only $4$ of the $800$ summaries ($3$ for Qwen3-235B and $1$ for Kimi-K2), each of which still lands within $17\%$ of its target (Appendix~\ref{app:dvr_convergence}). Unlike refinement from open-ended natural-language feedback~\cite{wadhwa2024learning}, this narrow feedback channel limits opportunities for feedback-induced drift.

Re-running the four judges on the length-controlled summaries collapses the verbosity channel. The longer summary now wins only $\approx\!57\%$ of decided pairs (per-judge $54.7$--$59.1\%$), against $\approx\!74\%$ on the uncontrolled drafts, and the length-preference correlation falls from $r_{pb}\approx+0.25$ to $r_{pb}\approx+0.16$ with the mean winner--loser gap shrinking from $\sim$13--14~words to $\sim$1~word. The residual preference signal can therefore be attributed more confidently to summary content than to verbosity, and we use these length-controlled summaries for the rest of our evaluations.

To assess whether refinement intensity is associated with judge outcomes, we correlate per-article DVR iteration counts with absolute preference scores and swap-flip counts. Mean Pearson correlations are $r=0.03$ across $32$ cells and $r=0.00$ across $64$ cells, respectively ($N=200$ per cell; Appendix~\ref{app:dvr_convergence}).

\section{Position bias}
\label{sec:position_bias}

Prior work has shown that LLMs exhibit a U-shaped positional pattern, with performance highest for information at the beginning and end of the context~\cite{liu2023lost} and causal attention biased toward earlier positions~\cite{wu2025emergence}. LLM judges likewise show position preferences in pairwise comparison, most often for the first candidate~\cite{wang2024large,zheng2023judging}.
We also observe a clear position bias. Across the 4 judges, each evaluating $1{,}200$ swap-pairs ($2{,}400$ ordered evaluations), the swap-flip rate -- the fraction of swap-pairs on which both orderings select the same slot, so that swapping the presentation order flips the judge's model-level verdict -- varies widely across judges and placements, under forced choice from $24\%$ (GPT-OSS-120b) up to $53\%$ (DeepSeek-V3.1) (Table~\ref{tab:master_bias}, \emph{preference} rows, \emph{No-Eq.}\ columns). Figure~\ref{fig:framework}(b) gives the full verdict-pair taxonomy this rate is computed from.

\paragraph{Terminology and outcome rates.} An \emph{Equal option} is a response option offered by the judging protocol. An \emph{Equal verdict} is its selection on one call. At the swap-pair level, \emph{stable abstention} means Equal in both orders, whereas \emph{tie/preference disagreement} means Equal in exactly one order. We use \emph{Any-Equal pairs} for their union when reporting the similarity analysis. Let $B$ count strict preference reversals (the same slot chosen twice), $U$ pairs with tie/preference disagreement, $T$ pairs with stable abstention, and $C$ consistent winners. For $N$ valid swap-pairs, strict reversal is $B/N$, total order disagreement is $(B+U)/N$, winner coverage is $C/N$, and stable abstention is $T/N$. Overall order consistency is $(C+T)/N$, counting both consistent
A/B preferences and stable Equal judgments. The Any-Equal pair share is $(T+U)/N$, and the share of ordered calls returning an Equal verdict is $(2T+U)/(2N)$.

\subsection{Characteristics of samples with position bias}
\label{sec:bias_similarity}

We characterize each evaluated pair using surface similarity, measured by cosine over OpenAI \texttt{text-embedding-3-large} document embeddings, and structural similarity, measured by the document-level Smatch F-score~\citep{cai-knight-2013-smatch} between the Abstract Meaning Representation (AMR) graphs produced by amrlib's BART-large \texttt{parse\_xfm} parser~\cite{jascob2022amrlib}. AMR captures predicate--argument structure independently of surface wording, providing a structural counterpart to the preference data. Given parser noise and Smatch's insensitivity to some semantic distinctions, we use it as a complementary signal rather than ground truth. We then compare the biased pairs against the consistent pairs from the same judge (those for which the two orderings agree on the same model rather than the same slot).

Across the four judges run on the full set (200~articles, 6~model pairs, both placements), biased and consistent pairs are essentially indistinguishable on both axes. On embedding cosine the two classes differ by less than $0.01$ (biased $0.876$ vs.\ consistent $0.868$, pooled and in each placement separately). Although this gap is statistically detectable given the sample size ($N\!=\!3{,}412$ biased, $6{,}188$ consistent; $p\!<\!10^{-6}$), its effect size is negligible (Cohen's $d\!\approx\!0.11$), and its direction is not even consistent across cells (biased are marginally more similar in only 5 of the 8 judge--placement cells). Smatch separates them no better (biased $0.375$, consistent $0.371$; $p=0.04$, $d=0.04$). Per-judge and per-placement breakdowns are given in Appendix~\ref{app:bias_similarity} (Table~\ref{tab:bias_similarity}).

Both classes sit in the same high-similarity regime (cosine $\sim\!0.86$--$0.87$), well above what would be expected for unrelated summaries. The summaries in every evaluated pair are near-paraphrastic, so surface and structural similarity do not explain when a judge exhibits position bias -- the flips occur on pairs only negligibly more similar than those the judge ranks consistently, by a margin far below the pooled separation between pairs exhibiting stable abstention and decided pairs (Section~\ref{sec:equal_option_mitigation}). We therefore read position bias not as a property of some hard subset of pairs but as a low-signal fallback. Because the length-controlled design leaves almost no quality signal to separate the candidates, a forced binary verdict lets position substitute for quality. This mechanism -- rather than any similarity-selected subset -- is what our mitigations target, by offering an Equal option (Section~\ref{sec:equal_option_mitigation}) or by widening the discriminative gap before the choice (Section~\ref{sec:reflection}).

\subsection{Order bias and intransitivity per judge}
\label{sec:cycle_rate}
The aggregate swap-flip rate tells us how often a judge is positionally biased, but not in which direction, and not whether the verdicts that survive the swap form a coherent ranking. We answer both questions by following the EigenBench protocol of \citet{chang2026eigenbenchcomparativebehavioralmeasure} (their Appendix~J), which decomposes position bias into directional \emph{primacy} and \emph{recency} components and adds an \emph{intransitivity} (cycle) check on the implied preference graph.

\paragraph{Directional decomposition.} For a judge $i$, an unordered summarizer pair and an article, we show the pair twice -- once in each order -- and record which slot won each time. If the judge names the same model both times, its preference is order-independent. If it picks the first slot both times, it has preferred whichever summary came first (\emph{primacy}, rate $O_{i,1}$), and if it picks the second slot both times, whatever came last (\emph{recency}, rate $O_{i,2}$). Both rates are fractions of unordered (pair, article) entries, so that $O_{i,1}+O_{i,2}$ recovers total order disagreement in the forced-choice preference rows of Table~\ref{tab:master_bias}. The formal definition is given in Appendix~\ref{app:order_bias_def}. As Table~\ref{tab:judge_cycle_rates} shows, under summaries-first placement recency dominates primacy for every judge -- by $3.6\times$ for DeepSeek-V3.1, $3.5\times$ for GPT-4o-mini and $4.8\times$ for GPT-5-mini, with only GPT-OSS-120b close to balanced. That direction is not a fixed property of the judge, however, since under article-first three of the four invert, GPT-OSS-120b and DeepSeek-V3.1 becoming primacy-dominated by $3.3\times$ and $52.2\times$. What holds instead is article-adjacency. The placement fixes which slot abuts the article -- the second under summaries-first, the first under article-first -- and that slot carries the larger rate in seven of the eight rows, all but GPT-4o-mini/AF. Closely matched summaries therefore leave the judge settling not on the one it read last but on the one it read next to the article. Although other studies report evaluator instability on closely matched summaries, with human assessors scoring identical summaries inconsistently~\citep{owczarzak2012assessing} and LLM judges struggling to separate candidates of similar quality~\citep{shen2023large}, the instability here is not symmetric noise but a direction set by prompt layout.

We also compare total order disagreement under article-first and summaries-first placement on the forced-choice preference set. No placement minimizes this rate across all judges. Article-first favors GPT-4o-mini and GPT-5-mini, whereas summaries-first favors DeepSeek~V3.1. GPT-OSS-120b is less placement-sensitive (Table~\ref{tab:master_bias}). We therefore report both throughout.

\paragraph{Intransitivity.} A judge could be order-independent on every individual pair and still be incoherent overall, by preferring $a$ over $b$, $b$ over $c$, and $c$ over $a$. We therefore turn from pairs to the $800$ article triples ($4$ per article), counting a triple as \emph{decided} when the judge names the same winning model in both orderings for all three of its constituent pairs -- the triples on which a ranking is even well-defined. The intransitivity rate $C_i$ is the share of decided triples containing a directed 3-cycle $a\!>\!b\!>\!c\!>\!a$. The rate $C_i^{\mathrm{abs}}$ recomputes it over all $800$ triples, a conservative lower bound that does not reward a judge for being too biased to rank anything.

Two things stand out. First, the decided counts are small -- $117$ to $161$ of $800$ in the worst cells -- and the three smallest belong to cells where the two directional rates are furthest apart. This is the order-bias result seen from the other side, in that position bias consumes most of the disagreement space before a ranking can be formed at all. Second, on the triples that do survive, preferences are essentially transitive ($C_i\!\leq\!1.7\%$, with zero cycles in four of the eight cells). The small denominators make $C_i$ a coherence check rather than a precise estimate -- a $0.00\%$ rate over $117$ triples still admits a true rate near $2.5\%$ -- but the residual signal passing the order-swap filter is internally consistent.

\begin{table}[h]
\centering
\footnotesize
\setlength{\tabcolsep}{3pt}
\begin{tabular}{llrrrrr}
\toprule
Judge & Plc. & $O_{i,1}$ & $O_{i,2}$ & Dec. & $C_i$ & $C_i^{\mathrm{abs}}$ \\
\midrule
DeepSeek-     & AF & \textbf{52.25} & 1.00           & 117 & 0.00 & 0.00 \\
V3.1          & SF & 7.17           & \textbf{25.50} & 274 & 0.73 & 0.25 \\
\addlinespace[2pt]
GPT-4o-       & AF & \textbf{10.67} & 17.33          & 321 & 0.00 & 0.00 \\
mini          & SF & 10.17          & \textbf{35.75} & 161 & 0.00 & 0.00 \\
\addlinespace[2pt]
GPT-5-        & AF & \textbf{15.42} & 11.67          & 315 & 0.32 & 0.12 \\
mini          & SF & 8.42           & \textbf{40.17} & 127 & 0.00 & 0.00 \\
\addlinespace[2pt]
GPT-OSS-      & AF & \textbf{19.25} & 5.92           & 352 & 1.70 & 0.75 \\
120b          & SF & 11.33          & \textbf{12.33} & 365 & 1.10 & 0.50 \\
\bottomrule
\end{tabular}
\caption{Per-judge order-bias and intransitivity under both placements, on the forced-choice preference set (AF = article-first, SF = summaries-first; $1{,}200$ unordered (pair, article) entries per row; \%\ except ``Dec.''). $O_{i,1}$ and $O_{i,2}$ are the primacy and recency rates. ``Dec.'' counts the decided article-triples (of $800$), those on which the judge names a consistent winner for all three constituent pairs in both orderings. $C_i$ is the cycle rate over those, and $C_i^{\mathrm{abs}}$ the same numerator divided by all $800$. \textbf{Bold} marks the article-adjacent slot's rate -- the first slot under AF, the second under SF -- larger in all rows but GPT-4o-mini/AF, though only by $1.0$ point for GPT-OSS-120b/SF.}
\label{tab:judge_cycle_rates}
\end{table}

\section{Position bias mitigation}
\label{sec:position_mitigation}

Length control removes the verbosity channel but leaves the positional one untouched. We therefore address position bias by offering an Equal option (Section~\ref{sec:equal_option_mitigation}) and asking the judge to reason about each criterion before returning a verdict (Section~\ref{sec:reflection}).

\subsection{Bias mitigation by offering an Equal option}
\label{sec:equal_option_mitigation}

An Equal option lets the judge abstain when it cannot distinguish the candidates, rather than forcing a potentially position-driven choice (Section~\ref{sec:bias_similarity}). To test whether this reduces position bias, we repeat the pairwise protocol with an explicit $\mathrm{Equal}$ option and strength $d=0$. The corresponding Equal-option contrast within per-criterion judging is shown horizontally in Figure~\ref{fig:framework}(c).

How often the Equal option is selected varies across judges evaluating the same data. Of the same $2{,}400$ ordered verdicts in every cell, the share returning an Equal verdict runs from $0\%$ (DeepSeek~V3.1) to $17\%$ (GPT-5-mini) under summaries-first, and from $0\%$ to $26\%$ under article-first (Table~\ref{tab:prompt_sensitivity}).

Pairs exhibiting stable abstention are more similar on average than pairs receiving A/B verdicts in both orders (the \emph{decided} group). Pooling both placements for GPT-4o-mini, GPT-5-mini and GPT-OSS-120b, the $404$ stable-abstention observations have mean embedding cosine $0.893$ versus $0.865$ for $5{,}557$ decided observations (Cohen's $d=0.38$). The broader Any-Equal group has $1{,}643$ observations and retains the same ordering ($0.890$ vs.\ $0.865$, $d=0.35$). With the same judges and pooling, AMR Smatch is $0.400$ versus $0.369$ for stable abstention ($d=0.31$), compared with $0.385$ versus $0.369$ for Any-Equal ($d=0.16$). Thus, the pooled separation persists under both definitions, with larger pooled effects for stable abstention. The groups overlap and do not constitute independent replications. The evidence varies across settings. Stable-abstention cosine differences have nominal Welch $p<10^{-3}$ in four settings, but $p=0.074$ for GPT-OSS-120b/summaries-first, while GPT-4o-mini/summaries-first has only one stable-abstention observation. DeepSeek~V3.1 has none. Appendix~\ref{app:equal_similarity} reports both definitions against the same decided group (Table~\ref{tab:equal_vs_decided_similarity}) and qualifies the tests. For calibration, the biased-vs-consistent contrast of Section~\ref{sec:bias_similarity} reaches only $d\approx0.11$ on cosine and $d=0.04$ on Smatch. These similarity diagnostics describe judge outcomes. They do not establish that the summaries are equivalent in quality.

The \emph{preference} rows of Table~\ref{tab:master_bias} show that offering Equal alone does not reliably improve order consistency under vanilla overall-preference judging: total order disagreement increases in six of the eight judge--placement settings. For GPT-5-mini/article-first, it rises from $27.1\%$ to $39.2\%$, even though strict reversals fall to $12.8\%$, because another $26.4\%$ of pairs exhibit tie/preference disagreement. Strict reversals decrease in seven of eight settings, but that reduction can be offset by preference-to-Equal changes (Appendix~\ref{app:coverage}). DeepSeek~V3.1 almost never uses Equal ($1$ of $1{,}200$ swap-pairs under article-first and none under summaries-first), yet its article-first disagreement rises from $53.2\%$ to $69.8\%$. Offering the option can therefore change A/B choices even when it is not selected; DeepSeek's first-slot share shifts from $75.6\%$ to $84.5\%$ (Appendix~\ref{app:prompt_sensitivity}).

%We nonetheless adopt the $\mathrm{Equal}$-enabled prompt for all subsequent pairwise analyses, since the alternative -- a binary prompt whose verdicts are dominated by positional artifacts -- is strictly worse.

\subsection{Bias mitigation through reflection}\label{sec:reflection}

\paragraph{Per-criterion judging.}

We next evaluate consistency, informativeness, coherence, and fluency separately using the EigenBench protocol~\cite{chang2026eigenbenchcomparativebehavioralmeasure}. Appendix~\ref{app:constitution} gives the constitution and prompts. Comparing the \emph{preference} and \emph{EB, no refl.}\ rows of Table~\ref{tab:master_bias}, with Equal availability held fixed, assesses this change from overall-preference to per-criterion judging without reflection. Under forced choice, total disagreement increases in seven of eight judge--placement settings; with Equal enabled, it decreases in five of eight.

\begin{table*}[!t]
\footnotesize
\setlength{\tabcolsep}{1pt}
\begin{tabular*}{\textwidth}{@{\extracolsep{\fill}} ll rrr rrr}
\toprule
 & & \multicolumn{3}{c}{Article-first} & \multicolumn{3}{c}{Summaries-first} \\
\cmidrule(lr){3-5}\cmidrule(lr){6-8}
 & & \multicolumn{2}{c}{Equal} & No-Eq. & \multicolumn{2}{c}{Equal} & No-Eq. \\
\cmidrule(lr){3-4}\cmidrule(lr){6-7}
Judge & Protocol & Disagr. & Cov. & Disagr. & Disagr. & Cov. & Disagr. \\
\midrule
GPT-4o-mini & pref. & 37.7 & 57.1 & 28.0 & 42.8 & 57.1 & 45.9 \\
 & EB, no refl. & 44.7 & 18.8 & 73.8 & 61.1 & 20.2 & 82.2 \\
 & EB, refl. & 36.0 & 13.4 & 61.6 & 36.4 & 8.9 & 63.2 \\
\addlinespace[2pt]
GPT-5-mini & pref. & 39.2 & 48.2 & 27.1 & 57.2 & 37.9 & 48.6 \\
 & EB, no refl. & 32.1 & 12.0 & 40.3 & 52.8 & 9.2 & 49.9 \\
 & EB, refl. & 18.2 & 14.9 & 26.7 & 27.8 & 13.2 & 34.5 \\
\addlinespace[2pt]
GPT-OSS-120b & pref. & 38.6 & 53.3 & 25.2 & 31.4 & 65.8 & 23.7 \\
 & EB, no refl. & 19.1 & 22.5 & 32.8 & 21.8 & 22.4 & 34.1 \\
 & EB, refl. & 11.4 & 25.0 & 40.9 & 12.1 & 23.6 & 39.2 \\
\addlinespace[2pt]
DeepSeek-V3.1 & pref. & 69.8 & 30.2 & 53.2 & 30.8 & 69.2 & 32.7 \\
 & EB, no refl. & 35.6 & 31.4 & 40.3 & 48.0 & 31.0 & 37.5 \\
 & EB, refl. & 18.1 & 23.2 & 45.1 & 13.0 & 29.2 & 30.9 \\
\bottomrule
\end{tabular*}
\caption{Total order disagreement and winner coverage (\%) under vanilla pairwise preference (pref.), EigenBench without reflection (EB, no refl.), and EigenBench with reflection (EB, refl.). \emph{Disagr.} counts strict preference reversals plus tie/preference disagreement, divided by all valid swap-pairs; lower is better. \emph{Cov.} counts pairs on which the same summary wins in both orders. Stable abstention contributes to order consistency but not winner coverage. Under No-Eq., disagreement equals the strict-reversal rate and coverage is $100$ minus disagreement. Preference rows use $N=1{,}200$ swap-pairs; EB rows pool four criteria ($N=4{,}800$, except $N=4{,}797$ for DeepSeek-V3.1 no-reflection/Equal/AF and reflection/Equal/SF). The EB rows form the reflection-by-Equal factorial comparison. Combining both interventions yields the lowest disagreement in all eight judge--placement settings; this ordering also holds on their common valid items. Comparisons are descriptive single-pass results. Appendix~\ref{app:coverage} reports strict reversals, tie/preference disagreement, and tie-adjusted bias.}
\label{tab:master_bias}
\end{table*}

Within this per-criterion protocol, reflection adds a preliminary step: each judge first writes a structured per-criterion analysis of the candidate summaries, and the pairwise comparison is then performed conditionally on these reflections (Figure~\ref{fig:reflection}). Its effect is shown by the vertical contrast in Figure~\ref{fig:framework}(c). Since position bias is stronger when the quality gap between candidates is smaller~\cite{shi2025judging}, the reflection step aims to widen the apparent gap before the choice is made. The comparison prompt also offers an Equal option, so the judge need not discriminate between reflections it judges equivalent.
Table~\ref{tab:master_bias} reports the full factorial grid. Compared with per-criterion judging without either intervention, combining an Equal option with reflection reduces total order disagreement by 44.4--65.4\% across the eight settings. Under summaries-first placement, the rate falls from 49.9\% to 27.8\% for GPT-5-mini and from 82.2\% to 36.4\% for GPT-4o-mini. The corresponding strict-reversal rates remain 7.8\% and 9.3\%. Across the eight judge--placement settings, the combined protocol yields stable abstention on 50.6--66.9\% of valid criterion-level swap-pairs and a consistent A/B winner on 8.9--29.2\%. The remaining outcomes are tie/preference disagreement (8.6--27.1\%) and strict preference reversal (2.2--9.8\%). Stable abstention counts as order consistency, while winner coverage measures how often the protocol consistently selects one summary. With reflection but without the Equal option, winner coverage is 36.8--73.3\%. We therefore report coverage next to every \emph{Equal}-enabled total-disagreement rate in Table~\ref{tab:master_bias}, with the tie-adjusted bias and the per-criterion breakdown in Appendix~\ref{app:coverage} (Table~\ref{tab:winner_coverage}).

The per-criterion factorial comparison in Table~\ref{tab:master_bias} shows why both interventions matter. Adding Equal without reflection lowers total disagreement in six of eight settings, and adding reflection without Equal lowers it in five of eight. Adding either intervention when the other is already enabled lowers disagreement in all eight. For GPT-4o-mini/summaries-first, neither intervention yields $82.2\%$ disagreement, Equal alone $61.1\%$, reflection alone $63.2\%$, and both $36.4\%$. Equal alone can also worsen disagreement: GPT-5-mini/summaries-first rises from $49.9\%$ to $52.8\%$, despite a much lower strict-reversal rate. Their combination is therefore best in every tested per-criterion setting, while their individual contributions depend on the setting. The common-valid-item comparison preserves this ordering. These descriptive total-disagreement results are distinct from the strict-reversal noise-band analysis in Appendix~\ref{app:repeats}. Among pairs receiving A/B verdicts in both orders, enabling Equal lowers tie-adjusted bias in seven of eight reflection settings; GPT-5-mini/summaries-first is the exception ($34.5\%$ to $37.2\%$). Appendix~\ref{app:coverage} reports the component rates and Appendix~\ref{app:eigenbench_distribution} the ordered-verdict distributions.

\paragraph{Disagreement is highest on informativeness.} With both mitigations enabled, median total order disagreement across the eight judge--placement settings is $42.7\%$ for informativeness, $15.3\%$ for consistency, $10.3\%$ for coherence, and $4.0\%$ for fluency (Figure~\ref{fig:framework}(d)). The appendix separates strict reversals from tie/preference disagreement (Tables~\ref{tab:eigenbench_per_criterion} and~\ref{tab:winner_coverage}). 

Informativeness requires comparing which important facts each summary preserves. Matching summary length controls verbosity but does not equalize information coverage, so the judge must still resolve differences in content selection. Coherence and fluency instead concern organization and linguistic form; under these criteria, the judges frequently return Equal in both orders. For GPT-5-mini with reflection under summaries-first placement, strict-reversal rates are $20.3\%$ for informativeness and $10.8\%$ for consistency, compared with $0.2\%$ for coherence and $0.0\%$ for fluency. This pattern is consistent with position serving as a fallback when the judge commits to a preference despite difficulty resolving a content difference (Section~\ref{sec:bias_similarity}). GPT-4o-mini is the exception, retaining substantial strict-reversal rates on coherence and fluency.

\subsection{Residual disagreement between judges}
\label{sec:residual_disagreement}
On the \emph{Equal}-enabled vanilla preference protocol, with both orderings pooled into a single net verdict per head-to-head, agreement between judges remains limited, with inter-judge Cohen's $\kappa$ staying in the fair-to-moderate range ($\kappa \in [0.29, 0.52]$) across all six judge pairs, even though the four judges produce broadly similar aggregate rankings of the summarizers (Appendix~\ref{app:judge_agreement}, with summarizer-level scores in Appendix~\ref{app:absolute_scores}). Judges concur on the ordering of summarizers considerably better than on the individual decisions that produce it -- a gap that single-judge, single-ordering protocols hide entirely.

\section{Conclusion}

We studied how to compare summaries fairly when both summarizers and judges are LLMs, using 200 recent Guardian Science articles. Length strongly influences uncontrolled judge preferences, while position bias persists after length control. Its direction depends on prompt layout. The article-adjacent slot is favored in seven of eight judge--placement settings. Biased and consistent pairs differ negligibly on embedding cosine and AMR Smatch, while residual strict reversals concentrate on informativeness for most judges. These findings are consistent with position serving as a fallback when quality differences are difficult to distinguish.

Combining an Equal option with reflection yields the lowest observed total order disagreement in all eight settings of the factorial ablation. Either intervention alone can increase disagreement. Under Equal-enabled vanilla preference judging, inter-judge agreement on individual comparisons remains fair-to-moderate ($\kappa \in [0.29, 0.52]$). These results establish reduced order disagreement, not improved agreement with human preference, which remains untested. Within this setting, we recommend candidate-side length control with Divide--Verify--Refine~\cite{zhang2025divide} and order-swapped judging with both an Equal option and reflection, reporting total disagreement, winner coverage, and tie-adjusted bias together.

\section*{Limitations}

Our study has several limitations. First, the evaluation scope is deliberately restricted, as we focus on a single domain (Guardian Science news). The domain and recency choice is a methodological decision -- Guardian Science articles published after the training cutoffs of the summarizers and judges minimize the risk of training-data contamination on the candidate side, while still providing reference-quality journalistic prose. Whether the bias patterns and mitigation effects we report transfer to other domains and languages remains open. Second, we evaluate four open-weight summarizers and four LLM judges. Although this is enough to observe consistent positional and length effects, it is a small sample from which to draw broader conclusions about specific model families or to rule out judge-specific idiosyncrasies. Third, we use no human annotation. Swap-flip rates, order-bias direction, per-criterion localization and inter-judge agreement are self-consistency diagnostics that are valid without gold labels, but they do not show that offering an Equal option or using reflection improves agreement with human preference, factuality, or model selection, and we do not know whether human annotators would judge pairs exhibiting stable abstention to be equivalent. We likewise make no claim about which summarizer is better. Fourth, our AMR-based diagnostics depend on a single parser (amrlib BART-large, \citealp{jascob2022amrlib}) and on Smatch as a structural similarity measure. Parser errors and Smatch's known insensitivity to certain semantic distinctions add noise that limits the power of the structural null in Section~\ref{sec:bias_similarity}, which therefore bounds rather than rules out a biased-versus-consistent difference. Fifth, our length-control step certifies length compliance rather than semantic preservation. Repeated DVR revisions can change content, and the target ratio, tolerance, and maximum number of refinement rounds determine how often and how extensively a summary is revised. Since conclusions may be sensitive to these protocol choices, and DVR's own constraint satisfaction depends on the number of refinement rounds~\cite{zhang2025divide}, we treat DVR as length normalization rather than a quality guarantee, and verify separately that iteration count does not itself predict judge outcomes (Section~\ref{sec:verbosity_mitigation}). We evaluated summaries under a 10\% length budget in the news domain, so our recommendation is limited to this setting. Whether the observed bias patterns and mitigation effects generalize to other domains and settings remains untested. We see these limitations as motivating future work on multi-domain, multilingual, human-anchored evaluation of LLM judges under controlled length and placement.

\paragraph{Potential risks.} Our contribution is an evaluation methodology rather than a deployed summarization system, and all source articles are professionally edited Guardian Science news. We therefore expect the direct risk of harm from this work to be low. The most concrete risk we see is methodological misuse, in which practitioners treat LLM-as-a-judge preferences as a substitute for human evaluation. Under a single-judge, single-ordering protocol we measure swap-flip rates of up to $69.8\%$. Our mitigations reduce the positional artifact rather than eliminate it, and inter-judge Cohen's $\kappa$ stays fair-to-moderate (Appendix~\ref{app:judge_agreement}). Such scores should not be used for high-stakes model selection. To respect The Guardian's content license and avoid redistributing potentially hallucinated paraphrases of news, we do not release the article texts or the generated summary texts themselves (Appendix~\ref{app:licenses}). Only article identifiers, judge verdicts, prompts, and code are released.

\paragraph{Use of AI assistants.} AI coding assistants (GitHub Copilot) were used for code scaffolding of evaluation and plotting scripts and for light copy-editing (grammar, phrasing) of the manuscript. The authors take responsibility for all content in the manuscript.

\bibliography{references}

\appendix
\section{Appendix}

\subsection{Self-preference}\label{app:self_preference}
GPT-OSS-120b is the only model in our design that appears both as a summarizer and as a judge, so the (judge~$=$~GPT-OSS-120b, summarizer~$=$~GPT-OSS-120b) cell is the one place where a judge scores its own output~\cite{panickssery2024llm}. For every one of the $16$ judge~$\times$~summarizer cells we compute the summarizer's win rate over its three opponents, pooling both presentation orders and dropping \emph{Equal} verdicts from the denominator. The self-preference contrast is the self cell against the pooled estimate of the same summarizer from the other three judges.

We find no self-preference (Table~\ref{tab:self_preference}). Across both placements and both prompt variants the self-judge scores its own summaries within $1.5$ points of the panel consensus, in three of four conditions \emph{below} it, and no contrast approaches significance ($p \geq 0.43$). The control column shows this is not a power problem. Computing the identical leave-one-out contrast for all $16$ cells gives a mean absolute deviation of $5.5$--$6.0$ points and a maximum of $13.9$--$15.5$ points, so ordinary judge idiosyncrasy is an order of magnitude larger than anything the self cell shows. The self cell is in fact the smallest or near-smallest of the $16$ deviations in every condition (rank $14$--$16$ of $16$ by magnitude). The largest deviations belong to cross cells instead -- GPT-4o-mini rates Qwen3-235B $+13.4$ to $+15.5$ points above consensus while GPT-OSS-120b rates the same model $-10.2$ to $-14.7$ points below it -- which is the same fair-to-moderate inter-judge disagreement quantified in Appendix~\ref{app:judge_agreement}, not a self-preference effect.

Self-preference could also affect the frequency of an Equal verdict rather than the win rate, if the self-judge were less willing to call a tie when its own summary is in the pair. It is not. On the \emph{Equal}-enabled set the self-judge's share of ordered calls returning an Equal verdict on pairs containing its own summaries exceeds its rate on the remaining pairs by $+3.1$ points (article-first) and $+0.8$ points (summaries-first) -- i.e.\ it returns an Equal verdict slightly \emph{more} often on calls involving its own summaries -- and GPT-4o-mini shows a comparable $+3.0$-point gap on the same pairs under article-first, indicating a property of those pairs rather than of the judge.

We report this as a null for our setting rather than as evidence against self-preference in general, because with a single self-judging model, length-matched candidates, and no human anchor, the design has the sensitivity to detect effects on the scale of the cross-judge disagreement we observe, but not smaller ones.

\begin{table}[h]
\small
\setlength{\tabcolsep}{3pt}
\begin{tabular*}{\columnwidth}{@{\extracolsep{\fill}} ll rr rr r}
\toprule
Prompt & Plc. & Self & Others & $\Delta$ & $p$ & Ctrl. \\
\midrule
Forced      & AF & 55.1 & 55.3 & $-0.2$ & 0.91 & 14.8 \\
choice      & SF & 53.7 & 54.8 & $-1.1$ & 0.50 & 15.5 \\
\addlinespace[2pt]
\emph{Equal}- & AF & 52.1 & 53.5 & $-1.5$ & 0.43 & 13.9 \\
enabled       & SF & 53.0 & 52.9 & $+0.1$ & 0.94 & 14.7 \\
\bottomrule
\end{tabular*}
\caption{Self-preference contrast for GPT-OSS-120b, the one model that is both a summarizer and a judge (AF = article-first, SF = summaries-first). ``Self'' is its win rate (\%) over its three opponents as scored by itself. ``Others'' is the same win rate pooled over the other three judges. $\Delta$ is their difference in points, with $p$ obtained from a two-proportion $z$-test. Win rates pool both presentation orders and exclude \emph{Equal} verdicts from the denominator ($n=1{,}200$ self / $3{,}600$ others under forced choice; $939$--$1{,}082$ self / $3{,}123$--$3{,}392$ others once ties are removed). ``Ctrl.'' is the largest absolute deviation observed when the identical leave-one-out contrast is computed for all $16$ judge~$\times$~summarizer cells, i.e.\ the scale of ordinary cross-judge disagreement in the same units. The self cell is indistinguishable from consensus while cross cells deviate by up to $15$ points.}
\label{tab:self_preference}
\end{table}

\subsection{Prompts used in the study}

\subsubsection{Summary prompt}\label{app:basic_prompt}

\begin{prompt}
Write a summary of the article in at most 
\{\{words\}\} words. 
Please think step by step. 
Please include your final summary in 
\texttt{<summary>} HERE YOUR summary \texttt{</summary>}.
Article: \{\{text\}\}
\end{prompt}

\subsubsection{DVR refinement feedback}\label{app:dvr_prompt}

The verifier's feedback to the summarizer is numeric. It reports the draft's word count, the target, and the signed deviation $\Delta w$, and identifies no content to add or drop. We use the following two variants for over- and under-length drafts, respectively.

\begin{prompt}
Note: The real target length is \{\{target\}\} words (the initial \{\{adjusted\_target\}\}-word limit was just a starting guideline to prevent overgeneration). Your summary has \{\{actual\}\} words, but the target is \{\{target\}\} words. You exceeded by \{\{excess\}\} words. Rewrite it to be \{\{target\}\} words or fewer while keeping the most important information.
\end{prompt}

\begin{prompt}
Note: The real target length is \{\{target\}\} words (the initial \{\{adjusted\_target\}\}-word limit was just a starting guideline to prevent overgeneration). Your summary has \{\{actual\}\} words, but the target is \{\{target\}\} words. You can add \{\{shortage\}\} more words to include more important details while staying at \{\{target\}\} words.
\end{prompt}

\paragraph{Restart schedule.} Refinement proceeds for at most five generations in a conversation where the model sees its previous drafts and the feedback above. If the constraint still fails, the loop restarts from a fresh conversation whose \emph{stated} word limit is lowered to $80\%$ of the target, and to $60\%$ for every subsequent restart. The acceptance test always uses the true target $L$ and its $\pm10\%$ band, and the feedback message continues to quote $L$, as shown above. The reduction is a heuristic against over-generation, which our first drafts in fact rarely exhibit (Appendix~\ref{app:dvr_convergence}). It takes effect only for the minority of articles whose summaries still fail the constraint after five generations, and the summary retained is in every case one that satisfies the band against $L$, or --- for the four articles that reach the safety cap --- the closest draft produced.

\subsubsection{Pairwise preference prompt}\label{app:pairwise_prompt}

The vanilla pairwise judge is shown the two candidate summaries (labeled \emph{A} and \emph{B}) together with the source article, in the order determined by the placement condition (summaries-first or article-first), and returns a single JSON verdict. The system prompt fixes the output contract, and the user prompt appends the preference instruction and the requested JSON schema to the content block.

\paragraph{System prompt.}
\begin{prompt}
You are comparing two different summaries of the same text.\\[2pt]
IMPORTANT: Provide your response as a plain JSON object WITHOUT any explanations, markdown formatting or code blocks.\\[2pt]
Example of correct format:\\
\{"preferred": "A", "difference": 2\}
\end{prompt}

\paragraph{User prompt.}
\begin{prompt}
Original Article: \{\{article\}\}\\[4pt]
Summary A : \{\{summary\_A\}\}\\[2pt]
Summary B : \{\{summary\_B\}\}\\[2pt]
Which summary do you prefer? Compare them and rate the difference on a scale of 1 to 3. You MUST pick either "A" or "B".\\
- The magnitude indicates how much better (1=slightly preferred, 2=preferred, 3=significantly preferred)\\
- If they seem very close, still pick the one you marginally prefer with a difference of 1.
\\[4pt]
Think step by step.\\
IMPORTANT: Return ONLY a JSON object in this exact format:\\
\{"preferred": "A or B", "difference": number\}\\[2pt]
Do not include any markdown formatting, code blocks, or additional text.
\end{prompt}

Under the summaries-first placement, the content block is reordered so that \texttt{Summary A} and \texttt{Summary B} precede the \texttt{Original Article}. All other prompt content is unchanged.

\paragraph{\emph{Equal} ablation.} In the Equal-enabled variant (Section~\ref{sec:equal_option_mitigation}), the preference instruction adds the tie option and the JSON schema extends the verdict to \texttt{"A, B or Equal"}:
\begin{prompt}
Original Article: \{\{article\}\}\\[4pt]
Summary A : \{\{summary\_A\}\}\\[2pt]
Summary B : \{\{summary\_B\}\}\\[2pt]
Which summary do you prefer? Compare them and rate the difference on a scale of 1 to 3. If you like them equally, select "Equal".\\
- The magnitude indicates how much better (1=slightly preferred, 2=preferred, 3=significantly preferred).\\
- If they are equally preferred, the difference should be 0.\\[4pt]
Think step by step.\\
IMPORTANT: Return ONLY a JSON object in this exact format:\\
\{"preferred": "A, B or Equal", "difference": number\}\\[2pt]
Do not include any markdown formatting, code blocks, or additional text.
\end{prompt}

\subsubsection{Per-criterion constitution and judging prompts}\label{app:constitution}

The per-criterion protocol of Section~\ref{sec:reflection} uses the same four criteria with and without reflection. For the reflection-based bias mitigation (Section~\ref{sec:reflection} and the EigenBench protocol of \citet{chang2026eigenbenchcomparativebehavioralmeasure}), each judge first writes a structured per-criterion analysis of every candidate summary against the constitution below. The pairwise comparison is then performed per criterion, conditioned on both reflections. The criteria used in our experiments are \emph{informativeness}, \emph{consistency}, \emph{coherence} and \emph{fluency}.

\begin{figure}[h]
\centering
\resizebox{\columnwidth}{!}{\input{figures/reflection}}
\caption{The reflection-based (EigenBench-style) judging protocol of
Section~\ref{sec:reflection}, applied to the length-matched pairs produced by
Stage~1 of Figure~\ref{fig:framework}(a). The judge first reflects on each summary
in isolation against the constitution $C$, naming no winner, so the two reflections
are order-independent. The same pair of reflections is then reused in both
presentation orders of the per-criterion comparison, which also exposes an explicit
\emph{Equal} verdict. A criterion counts as positionally biased when the two orderings
select the same slot rather than the same model.}
\label{fig:reflection}
\end{figure}
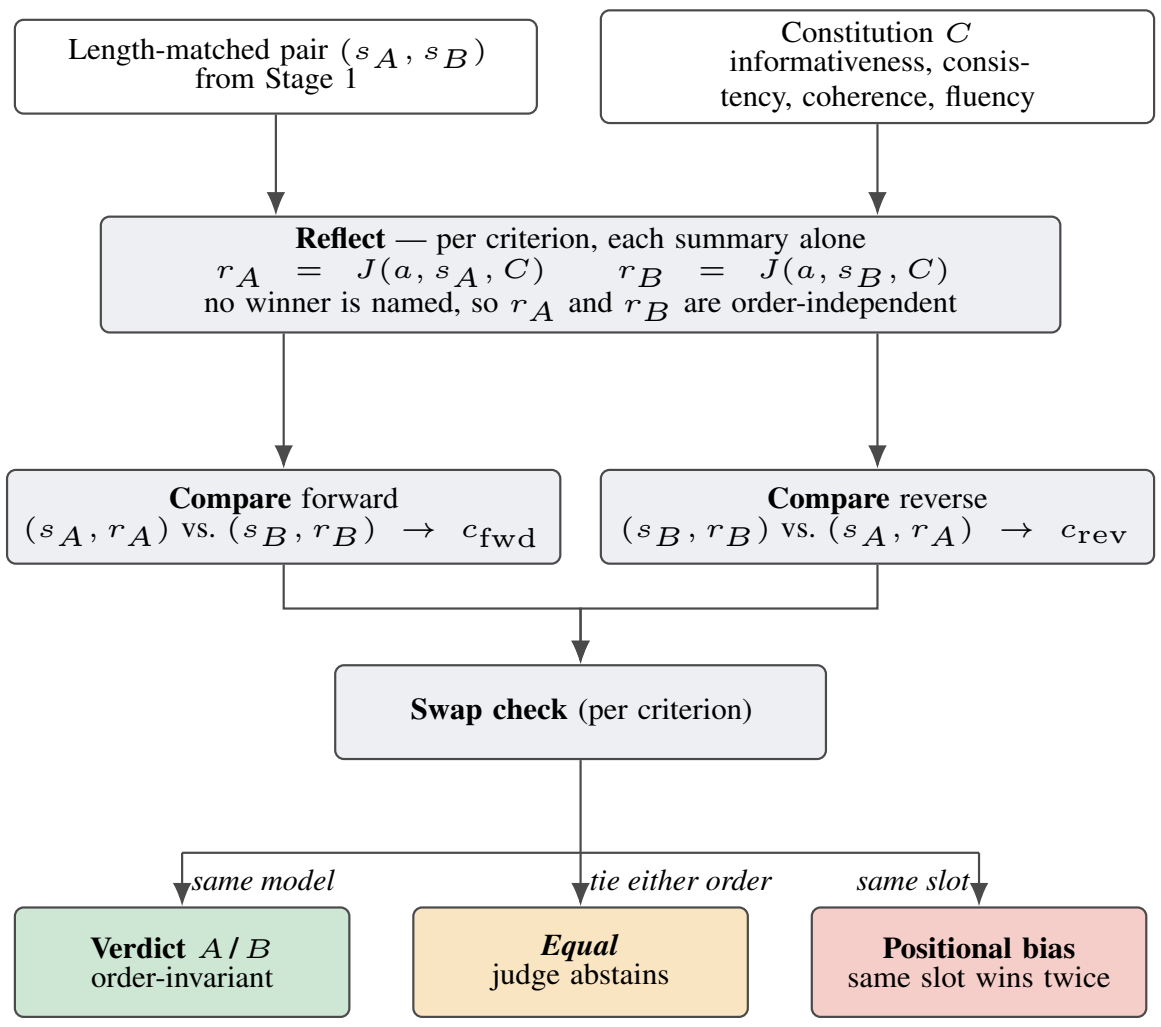

\paragraph{Reflection system prompt.}
\begin{prompt}
You are an impartial evaluator analyzing whether a SUMMARY aligns with a list of CRITERIA when judged against the original ARTICLE it was written for.\\[2pt]
You will be shown:\\
- a list of CRITERIA describing qualities a good summary should have,\\
- the SCENARIO (either the article followed by the summary, or the summary followed by the article, depending on the requested ordering),\\
- and you must produce a focused REFLECTION.\\[2pt]
Your reflection should:\\
1. Briefly restate, in your own words, what the article is about and what the summary claims.\\
2. For each criterion, comment on the extent to which the summary satisfies it, citing concrete evidence from the article and the summary. Be specific about strengths and weaknesses.\\
3. Do NOT pick a winner or assign a numeric score. The reflection is a structured analysis only; another step will perform the comparison.\\
4. Be concise: aim for one short paragraph per criterion.\\[2pt]
Wrap your final answer between \textless reflection\textgreater\ and \textless /reflection\textgreater\ tags. Do not output anything outside the tags.
\end{prompt}

\paragraph{Constitution (the \texttt{\textless criteria\textgreater} block, injected into both prompts).}
\begin{prompt}
1. Consistency: The factual alignment between the summary and the source article. A factually consistent summary contains only statements that are entailed by the source document, with no hallucinated facts, names, numbers, or relationships that are not supported by the article.\\[2pt]
2. Informativeness: How much of the important information and salient details from the source article are preserved in the summary. A more informative summary covers the article's main points, key facts, and meaningful context, rather than only superficial details.\\[2pt]
3. Coherence: The collective quality of all sentences in the summary. A coherent summary is well-structured and well-organized: it is not just a heap of related information but builds from sentence to sentence into a connected body of information about the topic.\\[2pt]
4. Fluency: The quality of the summary in terms of grammar, spelling, punctuation, word choice, and sentence structure. A fluent summary reads naturally, has no obvious grammatical errors, and uses clear, well-formed sentences.
\end{prompt}

\paragraph{Comparison system prompt (\emph{Equal}-enabled variant).}
\begin{prompt}
You are an impartial judge comparing two candidate SUMMARIES (the ``first response'' and the ``second response'') of the same source ARTICLE against a list of CRITERIA.\\[2pt]
You will be given:\\
- the list of CRITERIA (numbered 1..N),\\
- the SCENARIO (article + both summaries in the order specified),\\
- the first\_response (one summary),\\
- a first\_reflection produced by a separate evaluator about that summary,\\
- the second\_response (the other summary),\\
- a second\_reflection about that summary.\\[2pt]
For EACH criterion, decide which response is better, or whether they are tied. Use this output format, with one tag per criterion:\\
\textless criterion\_1\_choice\textgreater 0\textbar 1\textbar 2\textless /criterion\_1\_choice\textgreater\\
\textless criterion\_2\_choice\textgreater 0\textbar 1\textbar 2\textless /criterion\_2\_choice\textgreater\\
\ldots\\
\textless criterion\_N\_choice\textgreater 0\textbar 1\textbar 2\textless /criterion\_N\_choice\textgreater\\[2pt]
Where: 1 = the FIRST response is better, 2 = the SECOND response is better, 0 = TIE / equally good on this criterion.\\[2pt]
Rules: output exactly one choice tag per criterion, in order, using the literal tag names above; output ONLY the choice tags (no prose or markdown); judge each criterion independently; use ties (0) sparingly, only when the responses are genuinely indistinguishable.
\end{prompt}

In the no-\emph{Equal} ablation (the \emph{No-Eq.}\ columns of the EB rows in Table~\ref{tab:master_bias}) the comparison prompt is identical except that the Equal option is removed, so the value must be \texttt{1} or \texttt{2} for every criterion, and the judge is instructed that ties are not a valid choice (``if the responses look equivalent, pick whichever is marginally better'').

\paragraph{No-reflection ablation.} For Section~\ref{sec:reflection} (the \emph{EB, no refl.}\ rows of Table~\ref{tab:master_bias}), reflection generation is skipped and the two reflection blocks are omitted from the comparison input. The same constitution, article, summaries, and comparison system prompt are retained for the selected Equal setting. The reflection and no-reflection variants are crossed with both Equal settings in Section~\ref{sec:reflection}.

\subsection{DVR convergence and length-compliance details}
\label{app:dvr_convergence}

Table~\ref{tab:avg_iterations} pairs the first-draft compliance rates quoted in Section~\ref{sec:verbosity_mitigation} with the mean number of DVR rounds each model needed to satisfy the constraint. The two measures are not monotonically related, as Qwen3-235B misses the band only half the time yet needs more rounds on average ($4.43$) than Kimi-K2 ($2.80$), which misses it far more often. The dominant first-attempt failure mode is under-generation rather than over-generation, with median first-draft length at 93\% of the target for GPT-OSS-120b (median error $-7\%$), 91\% for Qwen3-235B ($-9\%$), 77\% for Kimi-K2 ($-23\%$), and just 23\% for GLM-4.5-Air ($-77\%$). 

\begin{table}[h!]
\small
\setlength{\tabcolsep}{4pt}
\begin{tabular*}{\columnwidth}{@{\extracolsep{\fill}} lrr}
\toprule
Model & In band (\%) & Avg.\ iter. \\
\midrule
GPT-OSS-120b & 60 & 1.43 \\
Qwen3-235B   & 50 & 4.43 \\
Kimi-K2      & 20 & 2.80 \\
GLM-4.5-Air  &  7 & 5.11 \\
\bottomrule
\end{tabular*}
\caption{Length compliance and DVR cost per model on 200 Guardian Science articles with a 10\% target ratio. ``In band'' is the fraction (\%) of first drafts inside the $\pm10\%$ target band (before any refinement). ``Avg.\ iter.'' is the mean number of DVR trials required to satisfy the constraint.}
\label{tab:avg_iterations}
\end{table}

\begin{figure}[h]
\begin{center}
\includegraphics[width=0.95\linewidth]{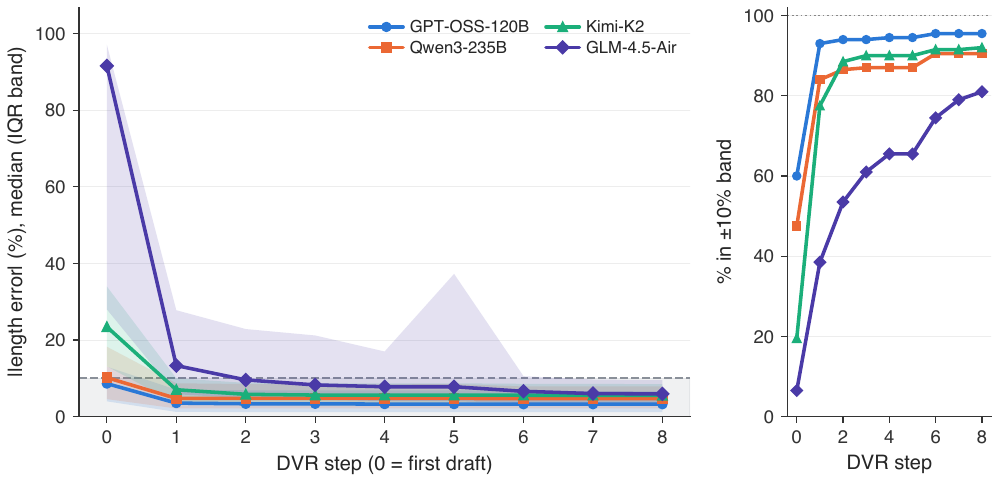}
\end{center}
\caption{DVR convergence across refinement iterations. The left panel shows the median absolute length error (\%) per model, with the IQR shaded. Step~0 is the first draft, and trajectories are carried forward after a model halts. The right panel shows the fraction of articles inside the $\pm10\%$ target band by step~$t$.}
\label{fig:dvr_convergence}
\end{figure}

Figure~\ref{fig:dvr_convergence} traces how the remaining models close that gap. GPT-OSS-120b reaches the target band almost immediately, with the median first draft already in band. Qwen3 and Kimi-K2 reach the band at the median by step~1, but Qwen3's mean iteration count is inflated by a heavy tail of unconverged trajectories. GLM-4.5-Air is slowest and reaches only $\sim\!80\%$ in-band coverage by step~8. Reaching the $100$-generation safety cap without satisfying the constraint is rare, happening for $4$ of the $800$ summaries ($3$ for Qwen3-235B and $1$ for Kimi-K2, with final deviations between $-16\%$ and $+14\%$ of target), and these are the only cases in which the retained summary is the closest draft rather than a compliant one. GPT-OSS-120b and GLM-4.5-Air never reach the cap, halting after at most $7$ and $35$ generations respectively.

\paragraph{Per-cell iteration-count correlations.} Section~\ref{sec:verbosity_mitigation} reports that DVR iteration count predicts neither judge preference nor position bias. The null holds cell by cell, not only on average. For preference, the $32$ judge~$\times$~summarizer~$\times$~placement coefficients ($N=200$ each) span $[-0.17, 0.18]$ with no cell reaching $|r|>0.2$, and their signs vary across judges rather than pointing consistently in one direction. For position bias, the $64$ coefficients (both protocols $\times$ both placements) span $[-0.10, 0.14]$. No individual cell therefore carries an association that the averaging could have masked.

\subsection{Biased vs.\ consistent pair similarity}\label{app:bias_similarity}
Table~\ref{tab:bias_similarity} gives the per-judge, per-placement breakdown behind the null reported in Section~\ref{sec:bias_similarity}, namely that neither surface nor structural similarity separates the pairs a judge flips on from those it ranks consistently.

\begin{table}[h]
\small
\setlength{\tabcolsep}{2.5pt}
\begin{tabular*}{\columnwidth}{@{\extracolsep{\fill}} ll rrr rrr}
\toprule
 & & \multicolumn{3}{c}{Biased} & \multicolumn{3}{c}{Consistent} \\
\cmidrule(lr){3-5}\cmidrule(lr){6-8}
Evaluator & Plc. & $N$ & Cos. & Sm. & $N$ & Cos. & Sm. \\
\midrule
DeepSeek-V3.1 & AF & 639 & 0.880 & 0.371 & 561 & 0.860 & 0.376 \\
              & SF & 392 & 0.869 & 0.379 & 808 & 0.872 & 0.370 \\
GPT-4o-mini   & AF & 336 & 0.869 & 0.379 & 864 & 0.872 & 0.371 \\
              & SF & 551 & 0.880 & 0.378 & 649 & 0.863 & 0.369 \\
GPT-5-mini    & AF & 325 & 0.874 & 0.372 & 875 & 0.870 & 0.374 \\
              & SF & 583 & 0.880 & 0.376 & 617 & 0.862 & 0.370 \\
GPT-OSS-120b  & AF & 302 & 0.876 & 0.374 & 898 & 0.869 & 0.373 \\
              & SF & 284 & 0.869 & 0.377 & 916 & 0.871 & 0.372 \\
\bottomrule
\end{tabular*}
\caption{Mean pairwise similarity per evaluator and placement (AF = article-first, SF = summaries-first) on \emph{biased} vs.\ \emph{consistent} pairs (200~articles, 6~model pairs), on the forced-choice A/B set (no \emph{Equal} option), so every swap-pair is either biased or consistent. Biased = both orderings pick the same slot. Consistent = both orderings pick the same model. Cos.\ is over OpenAI \texttt{text-embedding-3-large} embeddings. Sm.\ is document-level AMR Smatch. Biased and consistent pairs have negligible differences in effect size on both metrics (cosine difference $<\!0.01$, Cohen's $d\!\approx\!0.11$; Smatch $\approx\!0.37$ for both).}
\label{tab:bias_similarity}
\end{table}

\subsection{Order-bias decomposition}\label{app:order_bias_def}
Fix a judge $i$, an unordered summarizer pair $(j,k)$, and an article $l$. Let $r_{i,j,k,l}\in\{1,2\}$ be $1$ if the judge preferred whichever summary appeared first and $2$ if it preferred whichever appeared second. The primacy and recency order-bias rates used in Section~\ref{sec:cycle_rate} are the fractions of unordered (pair, article) entries on which the same slot won in both presentations:
\[
\textstyle
\begin{array}{r@{\;}c@{\;}l}
O_{i,1} &=& \frac{2}{|P_i|}\sum \mathbf{1}\!\left[r_{i,j,k,l}\!=\!r_{i,k,j,l}\!=\!1\right],\\[2pt]
O_{i,2} &=& \frac{2}{|P_i|}\sum \mathbf{1}\!\left[r_{i,j,k,l}\!=\!r_{i,k,j,l}\!=\!2\right],
\end{array}
\]
where $P_i$ is the set of ordered evaluations for judge $i$, and the factor $2/|P_i|$ normalizes by the number of unordered entries, since each contributes two ordered verdicts. By construction $O_{i,1}+O_{i,2}$ equals total order disagreement in the forced-choice preference rows of Table~\ref{tab:master_bias}.

\subsection{Prompt-sensitivity floor}\label{app:prompt_sensitivity}
Every cell of our grid is a single pass at temperature~$0$, so nothing in the reported rates bounds how far a cell would move under a change that carries no protocol meaning. We extract such a bound from data already collected. The forced-choice and \emph{Equal}-enabled preference runs place the same judge in front of the same $2{,}400$ ordered items under prompts that differ only in the tie option (Appendix~\ref{app:pairwise_prompt}). Restricting to the items the judge decides A/B under both isolates the verdict churn that the protocol difference does not explain.

Table~\ref{tab:prompt_sensitivity} reports it. Across the eight judge--placement cells, $10.2$--$24.4\%$ of decided verdicts change between the two wordings. Differences of a few points between cells of Table~\ref{tab:strict_reversal_ablation} are therefore not interpretable, and we base no claim on them.

The churn is not symmetric noise. If each of the two ordered verdicts of a swap-pair flipped independently with probability $p$, a biased pair would stay biased when both or neither flip and a consistent pair would become biased when exactly one flips, giving a predicted rate
\[
B' = B\left[(1-p)^2+p^2\right] + (1-B)\left[2p(1-p)\right],
\]
which for any $p>0$ pulls the rate toward $50\%$. The prediction is directly testable only for DeepSeek-V3.1, the one judge whose \emph{Equal} usage is small enough ($\leq\!1$ of $1{,}200$ pairs) that no denominator shift intervenes. It fails in both directions. Under article-first the observed rate is $69.8\%$ against a predicted $52.0\%$, and under summaries-first $30.8\%$ against $40.7\%$. What moves instead is the judge's slot preference as a whole -- DeepSeek's share of first-slot verdicts goes from $75.6\%$ to $84.5\%$ under article-first, and GPT-4o-mini's from $37.2\%$ to $59.5\%$ under summaries-first. For some judges, adding the tie option also changes the relative frequency of the two A/B verdicts. Separating this prompt sensitivity from decoding nondeterminism requires repeated passes under an unchanged prompt, as reported in Appendix~\ref{app:repeats}.

\begin{table}[h]
\small
\setlength{\tabcolsep}{3pt}
\begin{tabular*}{\columnwidth}{@{\extracolsep{\fill}} ll rr r}
\toprule
Judge & Plc. & Changed & Churn & Equal \\
\midrule
GPT-4o-mini   & AF & 21.4 & 10.6 & 12.1 \\
              & SF & 24.7 & 24.4 &  0.5 \\
GPT-5-mini    & AF & 33.3 & 10.2 & 25.8 \\
              & SF & 29.6 & 14.8 & 17.3 \\
GPT-OSS-120b  & AF & 31.2 & 13.8 & 20.2 \\
              & SF & 24.5 & 16.7 &  9.4 \\
DeepSeek-V3.1 & AF & 11.0 & 11.0 &  0.0 \\
              & SF & 13.3 & 13.3 &  0.0 \\
\bottomrule
\end{tabular*}
\caption{Verdict churn between the forced-choice and \emph{Equal}-enabled preference runs, which differ only in the tie option, per judge and placement (AF = article-first, SF = summaries-first). ``Changed'' counts the verdicts that differ between the two runs, including those that became \emph{Equal}, as a \%\ of the $2{,}400$ shared ordered items. ``Equal'' is the share of those same $2{,}400$ items routed to \emph{Equal}. ``Churn'' is the changed count restricted to items decided A/B in both runs, i.e.\ changes the tie option cannot explain, as a \%\ of those decided items rather than of all $2{,}400$. Churn of $10$--$24\%$ at temperature~$0$ sets the floor below which cell-level differences are not interpretable.}
\label{tab:prompt_sensitivity}
\end{table}

\subsection{Call-to-call stability}\label{app:repeats}
Appendix~\ref{app:prompt_sensitivity} bounds how far a cell moves under a prompt change. It cannot say how far a cell moves when nothing changes. We therefore re-issued identical prompts in three independent passes for every judge under summaries-first, with and without the \emph{Equal} option, on a fixed random subset of $50$ articles -- all six model pairs in both orders, i.e.\ $600$ ordered verdicts and $300$ swap-pairs per pass, $14{,}400$ verdicts in total, none of which failed. Decoding settings are those of the main grid, and the model version the endpoint reported serving was recorded for every pass. The provider that produced the released DeepSeek-V3.1 verdicts no longer serves the model without a dedicated deployment, so its passes ran on a different FP8 deployment, pinned to a single upstream so that provider routing could not masquerade as noise.

Table~\ref{tab:repeat_stability} reports the result. Churn between identical passes separates the judges into two groups. For the two non-reasoning judges it is small -- $1.6$--$2.6\%$ of decided verdicts for GPT-4o-mini and $4.1$--$4.3\%$ for DeepSeek-V3.1, against prompt-sensitivity floors of $24.4\%$ and $13.3\%$ -- so their floor measures sensitivity to the wording itself rather than sampling noise, which is the reading Appendix~\ref{app:prompt_sensitivity} gives it. For the two reasoning judges it is of the same order as the floor -- $11.2$--$12.8\%$ for GPT-5-mini against $14.8\%$, and $12.0$--$14.9\%$ for GPT-OSS-120b against $16.7\%$ -- so for them most of the floor is call-to-call variation. At temperature~$0$ these two judges are not deterministic, since the API exposes no temperature for GPT-5-mini, and GPT-OSS-120b samples during its reasoning phase.

The strict-reversal rates are far more stable than the verdicts that make them up. Across passes they span $1.3$--$3.3$ points for the non-reasoning judges and $3.0$--$5.7$ for the reasoning judges, and the released full-corpus rates of Table~\ref{tab:strict_reversal_ablation} lie inside these ranges or within about two points of them. One subset-restricted value falls outside. On the $50$ repeated articles the released pass for GPT-OSS-120b with \emph{Equal} gives $13.0\%$ against $17.7$--$20.7\%$ in the repeats. Only $0.64\%$ of random $50$-article subsets of that pass score as low, and its full-corpus value, $18.1\%$, sits $0.4$ points above the repeat range, so we read the subset value as sampling within a single pass rather than as a change in the judge. Every qualitative pattern replicates in every pass, as \emph{Equal} lowers the strict-reversal rate for three judges, and DeepSeek-V3.1 never selects it ($0$ of $1{,}800$ verdicts in which it was offered).

\paragraph{Stability of total order disagreement.} We also recompute total order disagreement from the same repeated verdicts, counting strict reversals and Equal in exactly one order (Table~\ref{tab:repeat_total_disagreement}). With Equal enabled, its observed ranges across the three passes are $39.7$--$43.0\%$ for GPT-4o-mini, $57.3$--$58.7\%$ for GPT-5-mini, $29.3$--$31.3\%$ for GPT-OSS-120b, and $31.7$--$33.0\%$ for DeepSeek-V3.1. The corresponding maximum-minus-minimum spreads are $3.3$, $1.3$, $2.0$, and $1.3$ percentage points. Without Equal, total disagreement equals strict reversal in every pass. These are observed three-pass ranges for the vanilla preference protocol under summaries-first placement, not confidence intervals. Article-first and the per-criterion/reflection protocols were not repeated, so their repeat variability is not measured here.

\paragraph{Reasoning effort.} The GPT-OSS-120b setting of Section~\ref{sec:setup} is the provider's default, which our requests do not pin, so we checked it with one further \emph{Equal}-enabled pass on the same subset at each explicit setting. The pass at \texttt{medium} differs from the three default passes by $13.1\%$ of decided verdicts, the same as they differ among themselves ($12.0\%$), whereas \texttt{low} and \texttt{high} differ by $20.4\%$ and $22.9\%$. The default is therefore \texttt{medium}. The setting is not neutral, since the share of \emph{Equal} verdicts falls from $12.7\%$ at \texttt{low} to $5.8\%$ at \texttt{medium} and $2.3\%$ at \texttt{high} on identical pairs, so how readily a judge returns an Equal verdict is a property of its decoding configuration as well as of the model -- a single pass per setting, which we report as an observation rather than a result.

\paragraph{Noise band.} We take as the noise band of a judge--placement cell the larger of its prompt-sensitivity churn (Table~\ref{tab:prompt_sensitivity}) and the pass-to-pass spread of its strict-reversal rate, both in percentage points. The churn is the larger in every cell, so the band runs from $10.2$ to $24.4$ points, and it is conservative where it is applied to the article-first cells and to the per-criterion protocols, which were not repeated. We do not use the independent-flip propagation of Appendix~\ref{app:prompt_sensitivity} for this purpose, because it vanishes for rates near $50\%$ and the churn is known not to be independent. Table~\ref{tab:noise_contrasts} marks every single-factor contrast of Table~\ref{tab:strict_reversal_ablation} against this band, and $30$ of the $56$ contrasts exceed it. All $16$ contrasts that remove the \emph{Equal} option under the per-criterion protocols do, as do $3$ of the $8$ on the plain preference protocol. Of the $16$ contrasts that add reflection, $6$ do and $10$ fall within the band. These noise-band comparisons concern strict preference reversals. Total-disagreement repeat ranges are reported separately in Table~\ref{tab:repeat_total_disagreement}; the noise-band thresholds are not applied to that metric.

\begin{table}[h]
\footnotesize
\setlength{\tabcolsep}{1.5pt}
\begin{tabular*}{\columnwidth}{@{\extracolsep{\fill}} ll rr c rr}
\toprule
 & & & & \multicolumn{1}{c}{Reversal} & \multicolumn{2}{c}{Released} \\
\cmidrule(lr){6-7}
Judge & Tie & Churn & Floor & range & 50 & 200 \\
\midrule
GPT-4o-mini   & No-Eq. &  1.6 & 24.4 & 45.3--46.7 & 47.0 & 45.9 \\
              & Equal  &  2.6 & 24.4 & 39.3--42.7 & 40.3 & 42.1 \\
GPT-5-mini    & No-Eq. & 12.8 & 14.8 & 42.3--46.3 & 45.7 & 48.6 \\
              & Equal  & 11.2 & 14.8 & 31.3--37.0 & 32.3 & 32.3 \\
GPT-OSS-120b  & No-Eq. & 14.9 & 16.7 & 20.7--26.0 & 21.0 & 23.7 \\
              & Equal  & 12.0 & 16.7 & 17.7--20.7 & 13.0 & 18.1 \\
DeepSeek-V3.1 & No-Eq. &  4.1 & 13.3 & 31.0--32.7 & 33.0 & 32.7 \\
              & Equal  &  4.3 & 13.3 & 31.7--33.0 & 34.0 & 30.8 \\
\bottomrule
\end{tabular*}
\caption{Three independent passes of the pairwise preference protocol on the same $50$ articles, summaries-first, with the tie option withheld (No-Eq.) and offered (Equal). ``Churn'' counts the ordered verdicts that differ between two passes, restricted to items decided A/B in both and averaged over the three pass pairs, as a \%\ of those items -- the quantity Table~\ref{tab:prompt_sensitivity} reports for the prompt change and whose summaries-first value is repeated here as ``Floor''. ``Reversal range'' gives the minimum and maximum strict-reversal rate (\%) over the three passes. ``Released'' is the strict-reversal rate of the single released pass of Table~\ref{tab:strict_reversal_ablation}, on the same $50$ articles and on all $200$.}
\label{tab:repeat_stability}
\end{table}

\begin{table}[t]
\footnotesize
\setlength{\tabcolsep}{2pt}
\begin{tabular*}{\columnwidth}{@{\extracolsep{\fill}} llrrr}
\toprule
Judge & Tie & \shortstack{Reversal\\range} & \shortstack{Disagr.\\range} & \shortstack{Disagr.\\spread} \\
\midrule
GPT-4o-mini & No-Eq. & 45.3--46.7 & 45.3--46.7 & 1.3 \\
 & Equal & 39.3--42.7 & 39.7--43.0 & 3.3 \\
GPT-5-mini & No-Eq. & 42.3--46.3 & 42.3--46.3 & 4.0 \\
 & Equal & 31.3--37.0 & 57.3--58.7 & 1.3 \\
GPT-OSS-120b & No-Eq. & 20.7--26.0 & 20.7--26.0 & 5.3 \\
 & Equal & 17.7--20.7 & 29.3--31.3 & 2.0 \\
DeepSeek-V3.1 & No-Eq. & 31.0--32.7 & 31.0--32.7 & 1.7 \\
 & Equal & 31.7--33.0 & 31.7--33.0 & 1.3 \\
\bottomrule
\end{tabular*}
\caption{Strict preference-reversal and total order-disagreement rates across three identical-prompt passes of vanilla pairwise preference, summaries-first, on the same $50$ articles ($300$ valid swap-pairs per pass). Ranges are minimum--maximum percentages; disagreement spread is maximum minus minimum in percentage points, computed before rounding. Total disagreement includes tie/preference disagreement and excludes stable abstention. No-Eq. and Equal use their respective fixed prompts. These ranges describe repeat variability in the measured conditions.}
\label{tab:repeat_total_disagreement}
\end{table}

\begin{table*}[t]
\small
\setlength{\tabcolsep}{4pt}
\begin{tabular*}{\textwidth}{@{\extracolsep{\fill}} ll r rrr rr rr}
\toprule
 & & & \multicolumn{3}{c}{\emph{Equal} $-$ No-Eq.} & \multicolumn{2}{c}{Protocol, No-Eq.} & \multicolumn{2}{c}{Protocol, \emph{Equal}} \\
\cmidrule(lr){4-6}\cmidrule(lr){7-8}\cmidrule(lr){9-10}
Judge & Plc. & Band & pref. & EB-nr & EB-r & EB-nr $-$ pref. & EB-r $-$ EB-nr & EB-nr $-$ pref. & EB-r $-$ EB-nr \\
\midrule
GPT-4o-mini   & AF & 10.6 & $(-4.0)$  & $-51.3$ & $-51.7$ & $+45.8$   & $-12.3$   & $(-1.5)$  & $-12.7$   \\
              & SF & 24.4 & $(-3.8)$  & $-43.2$ & $-53.9$ & $+36.3$   & $(-19.0)$ & $(-3.1)$  & $-29.7$   \\
GPT-5-mini    & AF & 10.2 & $-14.3$   & $-36.7$ & $-23.4$ & $+13.3$   & $-13.6$   & $(-9.1)$  & $(-0.3)$  \\
              & SF & 14.8 & $-16.2$   & $-34.9$ & $-26.7$ & $(+1.3)$  & $-15.4$   & $-17.4$   & $(-7.2)$  \\
GPT-OSS-120b  & AF & 13.8 & $(-10.8)$ & $-28.6$ & $-38.1$ & $(+7.6)$  & $(+8.1)$  & $(-10.1)$ & $(-1.4)$  \\
              & SF & 16.7 & $(-5.6)$  & $-28.7$ & $-36.7$ & $(+10.5)$ & $(+5.1)$  & $(-12.7)$ & $(-2.9)$  \\
DeepSeek-V3.1 & AF & 11.0 & $+16.5$   & $-31.0$ & $-38.9$ & $-13.0$   & $(+4.9)$  & $-60.4$   & $(-3.0)$  \\
              & SF & 13.3 & $(-1.9)$  & $-15.4$ & $-28.7$ & $(+4.9)$  & $(-6.6)$  & $(-8.6)$  & $-20.0$   \\
\bottomrule
\end{tabular*}
\caption{Single-factor contrasts of Table~\ref{tab:strict_reversal_ablation}, in percentage points of strict-reversal rate, against the noise band of each judge--placement cell (``Band'', the larger of the prompt-sensitivity churn and the pass-to-pass spread of Table~\ref{tab:repeat_stability}). Parenthesized differences do not exceed the band and are not interpreted. The first three columns add the \emph{Equal} option under each protocol (pref.\ = pairwise preference, EB-nr = EigenBench without reflection, EB-r = with reflection). The remaining four change the protocol with the tie setting held fixed.}
\label{tab:noise_contrasts}
\end{table*}

\subsection{Similarity of stable abstention, Any-Equal and decided pairs}\label{app:equal_similarity}
Table~\ref{tab:equal_vs_decided_similarity} compares stable abstention and the broader Any-Equal grouping with a shared decided group. Restricting the target group to Equal/Equal excludes mixed Equal/preference outcomes rather than moving them into the decided group. For the three judges with stable abstentions, this reduces the target count from $1{,}643$ to $404$, while the decided count remains $5{,}557$.

Under the original Welch-test convention, stable-abstention cosine has nominal $p<10^{-3}$ in four settings. GPT-OSS-120b/SF has $p=0.074$, and GPT-4o-mini/SF has only one observation. The Any-Equal cosine contrast has $p<10^{-3}$ in all six settings for these three judges. DeepSeek~V3.1 has no stable abstentions and only one Any-Equal pair (AF), whose cosine is below the decided mean. Neither DeepSeek cell supports an inferential comparison. Pooling both layouts of the three judges, stable-abstention cosine has $p=6.90\times10^{-20}$, while Any-Equal has $p=1.67\times10^{-47}$.

For per-sentence AMR Smatch with the same pooling, stable abstention gives $0.400$ vs.\ $0.369$ ($d=0.31$, nominal $p=5.79\times10^{-8}$), compared with $0.385$ vs.\ $0.369$ for Any-Equal ($d=0.16$, $p=1.80\times10^{-8}$). Stable-abstention Smatch effects vary by judge, with $d=0.56$ for GPT-4o-mini/AF ($p=0.0016$), $d=0.14$/$0.11$ for GPT-5-mini AF/SF ($p=0.171$/$0.355$), and $d=0.46$/$0.53$ for GPT-OSS-120b AF/SF ($p=1.15\times10^{-5}$/$0.0067$). GPT-4o-mini/SF again has only one observation. Including DeepSeek in the pooled decided group yields stable-abstention $d=0.35$ on cosine and $d=0.30$ on Smatch, preserving the direction.

These unadjusted Welch tests follow the original analysis. Shared articles and repeated summary pairs across judges and layouts violate independence. We therefore treat the pooled effects as descriptive and do not interpret the nominal $p$-values as cluster-adjusted evidence. Stable abstention and Any-Equal are overlapping comparisons, not independent confirmations.

\begin{table*}[t]
\small
\centering
\setlength{\tabcolsep}{4pt}
\begin{tabular*}{\textwidth}{@{\extracolsep{\fill}} ll rr rrr rrr}
\toprule
 & & \multicolumn{2}{c}{Decided} & \multicolumn{3}{c}{Any-Equal pairs} & \multicolumn{3}{c}{Stable abstention} \\
\cmidrule(lr){3-4}\cmidrule(lr){5-7}\cmidrule(lr){8-10}
Judge & Plc. & $N$ & Cos. & $N$ & Cos. & $d$ & $N$ & Cos. & $d$ \\
\midrule
GPT-4o-mini & AF & 973 & 0.865 & 227 & 0.896 & 0.45 & 63 & 0.900 & 0.48 \\
 & SF & 1190 & 0.870 & 10 & 0.922 & 0.73 & 1 & 0.912 & -- \\
GPT-5-mini & AF & 732 & 0.859 & 468 & 0.889 & 0.42 & 151 & 0.895 & 0.49 \\
 & SF & 843 & 0.862 & 357 & 0.890 & 0.40 & 59 & 0.889 & 0.34 \\
GPT-OSS-120b & AF & 812 & 0.864 & 388 & 0.885 & 0.30 & 97 & 0.888 & 0.32 \\
 & SF & 1007 & 0.867 & 193 & 0.891 & 0.34 & 33 & 0.887 & 0.28 \\
DeepSeek-V3.1 & AF & 1199 & 0.871 & 1 & 0.696 & -- & 0 & -- & -- \\
 & SF & 1200 & 0.871 & 0 & -- & -- & 0 & -- & -- \\
\midrule
\multicolumn{2}{l}{Pooled (excluding DeepSeek)} & 5557 & 0.865 & 1643 & 0.890 & 0.35 & 404 & 0.893 & 0.38 \\
\bottomrule
\end{tabular*}
\caption{Embedding similarity under two definitions of an Equal outcome, using the Equal-enabled preference prompt ($200$ articles, six model pairs, $1{,}200$ swap-pairs per judge and placement; AF = article-first, SF = summaries-first). \emph{Stable abstention} means Equal in both orders. \emph{Any-Equal} includes stable abstention and tie/preference disagreement. Both are compared with the same \emph{decided} group (A/B in both orders, whether biased or consistent). Cos.\ is mean OpenAI \texttt{text-embedding-3-large} cosine. Each $d$ compares its group with decided pairs. The stable group is a subset of Any-Equal, so their counts must not be added. Means are shown for singleton groups, but $d$ and tests are omitted when $N<2$. The pooled row combines both layouts of the three judges with stable abstentions. Cell-level and pooled test qualifications are in Appendix~\ref{app:equal_similarity}.}
\label{tab:equal_vs_decided_similarity}
\end{table*}

\subsection{EigenBench verdict distribution}\label{app:eigenbench_distribution}

Table~\ref{tab:eigenbench_choice_distribution} gives the ordered-verdict distribution for the per-criterion conditions in Table~\ref{tab:master_bias}; Table~\ref{tab:strict_reversal_ablation} reports the corresponding swap-pair disagreement components. With reflection the share of ordered calls returning an Equal verdict is high but stable across judges and placements ($63.2$--$74.3\%$ in every cell), so the share of ordered judgments that remain decided is comparable across the conditions we contrast. The corresponding swap-pair outcomes are reported in Appendix~\ref{app:coverage}. The residual first/second split is where the directional asymmetry of Section~\ref{sec:cycle_rate} becomes directly visible. It is near-balanced for GPT-OSS-120b and DeepSeek~V3.1 with reflection, and most lopsided without reflection under summaries-first, where the second slot takes $47.6\%$ of judgments against the first slot's $22.6\%$ for GPT-4o-mini and $47.8\%$ against $18.3\%$ for DeepSeek~V3.1.

\begin{table}[h]
\footnotesize
\setlength{\tabcolsep}{2pt}
\begin{tabular*}{\columnwidth}{@{\extracolsep{\fill}} llrrr}
\toprule
Judge & Condition & 1st wins & \shortstack{Equal\\verdict} & 2nd wins \\
\midrule
GPT-4o-mini   & refl, AF    & 11.5\% & 63.7\% & 24.8\% \\
              & refl, SF    &  7.3\% & 68.3\% & 24.4\% \\
              & no-refl, AF & 25.3\% & 47.6\% & 27.1\% \\
              & no-refl, SF & 22.6\% & 29.8\% & 47.6\% \\
GPT-5-mini    & refl, AF    &  8.9\% & 74.3\% & 16.8\% \\
              & refl, SF    &  7.3\% & 69.0\% & 23.7\% \\
              & no-refl, AF &  7.5\% & 70.2\% & 22.3\% \\
              & no-refl, SF &  5.4\% & 56.9\% & 37.7\% \\
GPT-OSS-120b  & refl, AF    & 16.9\% & 67.9\% & 15.2\% \\
              & refl, SF    & 15.0\% & 69.0\% & 15.9\% \\
              & no-refl, AF & 16.0\% & 65.9\% & 18.1\% \\
              & no-refl, SF & 15.8\% & 64.0\% & 20.2\% \\
DeepSeek-V3.1 & refl, AF    & 23.0\% & 64.6\% & 12.4\% \\
              & refl, SF    & 18.6\% & 63.2\% & 18.2\% \\
              & no-refl, AF & 34.5\% & 46.2\% & 19.3\% \\
              & no-refl, SF & 18.3\% & 33.9\% & 47.8\% \\
\bottomrule
\end{tabular*}
\caption{EigenBench verdict distribution per condition with the \emph{Equal} option allowed (forward and reverse orderings combined; $N = 9600$ per row), per placement (AF = article-first, SF = summaries-first). Columns give the share of per-criterion judgments assigned to the first slot, an Equal verdict, and the second slot.}
\label{tab:eigenbench_choice_distribution}
\end{table}

\subsection{Winner coverage and tie-adjusted bias}\label{app:coverage}

Neither stable abstention nor tie/preference disagreement contributes to the strict-reversal rate, although only stable abstention is order-consistent. A judge exhibiting stable abstention on every pair would have perfect order consistency and zero winner coverage. Table~\ref{tab:winner_coverage} therefore reports, for the \emph{Equal}-enabled vanilla preference and reflection conditions, how the swap-pairs divide among the four outcomes two order-swapped verdicts can take -- \emph{consistent A/B preference} (the same summary preferred in both orders), \emph{strict preference reversal} (the same slot in both orders), \emph{tie/preference disagreement} (\emph{Equal} in exactly one order) and \emph{stable abstention} (\emph{Equal} in both orders) -- and two derived rates, \emph{winner coverage}, the share of all pairs with a consistent A/B preference, and the \emph{tie-adjusted bias}, the strict-reversal share among pairs receiving A/B verdicts in both orders. Tie-adjusted bias is calculated only among pairs receiving A/B verdicts in both orders. Pairs exhibiting stable abstention or tie/preference disagreement are excluded. The \emph{Rev.}\ column is the strict-reversal rate of Table~\ref{tab:strict_reversal_ablation}. Under a forced-choice prompt, winner coverage is simply $100\%$ minus that rate. Total order disagreement in Table~\ref{tab:master_bias} is the sum of the \emph{Rev.}\ and Tie/preference disagreement shares, computed from unrounded counts. Equal in both orders is stable abstention. Equal in only one order is tie/preference disagreement, even though the pair yields no order-independent winner.

Three regularities stand out. First, under reflection with \emph{Equal} the pooled winner coverage is $8.9$--$29.2\%$, and the tie/preference disagreement share ($8.6$--$27.1\%$) is of the same order as the coverage itself, so the ordered Equal-verdict rates of Table~\ref{tab:eigenbench_choice_distribution} do not distinguish stable Equal judgments from tie/preference disagreement. Second, winner coverage varies by criterion. On fluency and coherence it is $1.5$--$11.2\%$ and $2.4$--$16.3\%$, with stable abstention on $59$--$98\%$ of pairs, whereas on informativeness it is $11.9$--$68.6\%$ -- judges exhibit stable abstention where they judge two length-matched summaries indistinguishable in both orders and return A/B preferences where they judge the summaries distinguishable, which is also where the residual bias concentrates (Section~\ref{sec:reflection}). Third, the tie-adjusted bias separates the judges. With reflection it is $6.9$--$21.3\%$ for DeepSeek~V3.1 and GPT-OSS-120b but $42.4$--$51.3\%$ for GPT-4o-mini, whose surviving verdicts follow the slot about as often as a coin would. All counts are recomputed from the stored verdicts (\texttt{scripts/compute\_decision\_coverage.py}).

\begin{table}[t]
\footnotesize
\setlength{\tabcolsep}{1pt}
\begin{tabular*}{\columnwidth}{@{\extracolsep{\fill}} ll rrr rrr}
\toprule
 & & \multicolumn{3}{c}{Article-first} & \multicolumn{3}{c}{Summaries-first} \\
\cmidrule(lr){3-5}\cmidrule(lr){6-8}
 & & \multicolumn{2}{c}{Equal} & No-Eq. & \multicolumn{2}{c}{Equal} & No-Eq. \\
\cmidrule(lr){3-4}\cmidrule(lr){6-7}
Judge & Protocol & Rev. & Mix. & Rev. & Rev. & Mix. & Rev. \\
\midrule
GPT-4o- & pref. & 24.0 & 13.7 & 28.0 & 42.1 & 0.8 & 45.9 \\
mini & EB, no refl. & 22.5 & 22.2 & 73.8 & 39.0 & 22.1 & 82.2 \\
 & EB, refl. & 9.8 & 26.2 & 61.6 & 9.3 & 27.1 & 63.2 \\
\addlinespace[2pt]
GPT-5- & pref. & 12.8 & 26.4 & 27.1 & 32.3 & 24.8 & 48.6 \\
mini & EB, no refl. & 3.6 & 28.4 & 40.3 & 15.0 & 37.8 & 49.9 \\
 & EB, refl. & 3.4 & 14.9 & 26.7 & 7.8 & 19.9 & 34.5 \\
\addlinespace[2pt]
GPT-OSS- & pref. & 14.3 & 24.2 & 25.2 & 18.1 & 13.3 & 23.7 \\
120b & EB, no refl. & 4.2 & 15.0 & 32.8 & 5.4 & 16.4 & 34.1 \\
 & EB, refl. & 2.8 & 8.6 & 40.9 & 2.5 & 9.5 & 39.2 \\
\addlinespace[2pt]
DeepSeek- & pref. & 69.8 & 0.1 & 53.2 & 30.8 & 0.0 & 32.7 \\
V3.1 & EB, no refl. & 9.3 & 26.3 & 40.3 & 22.1 & 25.8 & 37.5 \\
 & EB, refl. & 6.3 & 11.8 & 45.1 & 2.2 & 10.8 & 30.9 \\
\bottomrule
\end{tabular*}
\caption{Strict preference-reversal rates (\%) for the protocol grid of Table~\ref{tab:master_bias}. \emph{Rev.} counts the same presentation slot selected in both orders, divided by all valid swap-pairs. \emph{Mix.} counts Equal in exactly one order (tie/preference disagreement), also divided by all valid swap-pairs. Rev. plus Mix. gives the Equal-enabled disagreement columns of Table~\ref{tab:master_bias}, up to rounding. Under No-Eq., strict reversal equals total order disagreement; with Equal enabled, tie/preference disagreement is an additional component. Denominators and protocol abbreviations follow Table~\ref{tab:master_bias}. The prompt-sensitivity and noise-band analyses assess this strict-reversal metric; Tables~\ref{tab:repeat_stability} and~\ref{tab:repeat_total_disagreement} report repeat stability for both metrics.}
\label{tab:strict_reversal_ablation}
\end{table}

\begin{table}[h]
\small
\setlength{\tabcolsep}{3pt}
\begin{tabular*}{\columnwidth}{@{\extracolsep{\fill}} llrrrr}
\toprule
Judge & Plc. & Cons. & Info. & Coh. & Flu. \\
\midrule
GPT-4o-mini & AF & 2.8\% & 13.8\% & 9.3\% & 13.4\% \\
 & SF & 2.4\% & 20.7\% & 5.6\% & 8.6\% \\
GPT-5-mini & AF & 5.5\% & 7.8\% & 0.1\% & 0.1\% \\
 & SF & 10.8\% & 20.3\% & 0.2\% & 0.0\% \\
GPT-OSS-120b & AF & 0.1\% & 10.7\% & 0.2\% & 0.1\% \\
 & SF & 0.2\% & 9.8\% & 0.1\% & 0.1\% \\
DeepSeek-V3.1 & AF & 0.6\% & 24.3\% & 0.2\% & 0.0\% \\
 & SF & 1.2\% & 7.3\% & 0.2\% & 0.0\% \\
\bottomrule
\end{tabular*}
\caption{EigenBench strict preference-reversal rates broken down by criterion (consistency, informativeness, coherence, fluency) with reflection and the \emph{Equal} option enabled, per placement (AF = article-first, SF = summaries-first). Pooling valid swap-pairs across criteria gives the \emph{Rev.}\ entry in the corresponding \emph{Equal}/\emph{EB, refl.}\ row of Table~\ref{tab:strict_reversal_ablation}. Strict reversals concentrate on informativeness for most judges and are rare on fluency and coherence except for GPT-4o-mini. Informativeness leads without the reflection step too, though less sharply (Appendix~\ref{app:per_criterion_norefl}).}
\label{tab:eigenbench_per_criterion}
\end{table}

\begin{table*}[t]
\footnotesize
\setlength{\tabcolsep}{4pt}
\begin{tabular*}{\textwidth}{@{\extracolsep{\fill}} lllrrrrr}
\toprule
Judge & Plc. & Protocol / criterion & Winner coverage & Rev. & Rev.$\mid$dec. & \shortstack{Tie/preference\\disagreement} & \shortstack{Stable\\abstention} \\
\midrule
GPT-4o-mini    & AF & Preference, overall        & 57.1 & 24.0 & 29.6 & 13.7 & 5.2 \\
               &    & Reflection: consistency    & 10.8 & 2.8 & 20.9 & 16.5 & 69.9 \\
               &    & \quad informativeness      & 21.8 & 13.8 & 38.6 & 33.3 & 31.1 \\
               &    & \quad coherence            & 16.3 & 9.3 & 36.4 & 32.2 & 42.2 \\
               &    & \quad fluency              & 4.5 & 13.4 & 74.9 & 22.8 & 59.3 \\
               &    & Reflection, all criteria   & 13.4 & 9.8 & 42.4 & 26.2 & 50.6 \\
\addlinespace[1pt]
               & SF & Preference, overall        & 57.1 & 42.1 & 42.4 & 0.8 & 0.1 \\
               &    & Reflection: consistency    & 6.8 & 2.4 & 26.1 & 17.8 & 73.0 \\
               &    & \quad informativeness      & 11.9 & 20.7 & 63.4 & 40.2 & 27.2 \\
               &    & \quad coherence            & 11.6 & 5.6 & 32.5 & 29.2 & 53.6 \\
               &    & \quad fluency              & 5.1 & 8.6 & 62.8 & 21.3 & 65.0 \\
               &    & Reflection, all criteria   & 8.9 & 9.3 & 51.3 & 27.1 & 54.7 \\
\addlinespace[1pt]
\midrule
GPT-5-mini     & AF & Preference, overall        & 48.2 & 12.8 & 20.9 & 26.4 & 12.6 \\
               &    & Reflection: consistency    & 15.9 & 5.5 & 25.7 & 20.8 & 57.8 \\
               &    & \quad informativeness      & 39.7 & 7.8 & 16.3 & 34.2 & 18.3 \\
               &    & \quad coherence            & 2.5 & 0.1 & 3.2 & 3.7 & 93.8 \\
               &    & \quad fluency              & 1.5 & 0.1 & 5.3 & 0.8 & 97.6 \\
               &    & Reflection, all criteria   & 14.9 & 3.4 & 18.4 & 14.9 & 66.9 \\
\addlinespace[1pt]
               & SF & Preference, overall        & 37.9 & 32.3 & 46.0 & 24.8 & 4.9 \\
               &    & Reflection: consistency    & 14.7 & 10.8 & 42.3 & 35.1 & 39.5 \\
               &    & \quad informativeness      & 32.8 & 20.3 & 38.2 & 37.4 & 9.4 \\
               &    & \quad coherence            & 2.4 & 0.2 & 6.5 & 5.9 & 91.5 \\
               &    & \quad fluency              & 2.8 & 0.0 & 0.0 & 1.3 & 95.8 \\
               &    & Reflection, all criteria   & 13.2 & 7.8 & 37.2 & 19.9 & 59.1 \\
\addlinespace[1pt]
\midrule
GPT-OSS-120b   & AF & Preference, overall        & 53.3 & 14.3 & 21.2 & 24.2 & 8.1 \\
               &    & Reflection: consistency    & 15.3 & 0.1 & 0.5 & 4.0 & 80.6 \\
               &    & \quad informativeness      & 66.3 & 10.7 & 13.9 & 17.2 & 5.8 \\
               &    & \quad coherence            & 10.6 & 0.2 & 2.3 & 9.2 & 80.0 \\
               &    & \quad fluency              & 7.9 & 0.1 & 1.0 & 4.1 & 87.9 \\
               &    & Reflection, all criteria   & 25.0 & 2.8 & 10.0 & 8.6 & 63.6 \\
\addlinespace[1pt]
               & SF & Preference, overall        & 65.8 & 18.1 & 21.5 & 13.3 & 2.8 \\
               &    & Reflection: consistency    & 12.2 & 0.2 & 2.0 & 5.1 & 82.4 \\
               &    & \quad informativeness      & 64.9 & 9.8 & 13.1 & 18.2 & 7.2 \\
               &    & \quad coherence            & 10.6 & 0.1 & 0.8 & 10.1 & 79.2 \\
               &    & \quad fluency              & 6.8 & 0.1 & 1.2 & 4.8 & 88.2 \\
               &    & Reflection, all criteria   & 23.6 & 2.5 & 9.7 & 9.5 & 64.3 \\
\addlinespace[1pt]
\midrule
DeepSeek-V3.1  & AF & Preference, overall        & 30.2 & 69.8 & 69.8 & 0.1 & 0.0 \\
               &    & Reflection: consistency    & 15.9 & 0.6 & 3.5 & 7.2 & 76.3 \\
               &    & \quad informativeness      & 54.2 & 24.3 & 31.0 & 19.1 & 2.3 \\
               &    & \quad coherence            & 13.3 & 0.2 & 1.8 & 17.2 & 69.2 \\
               &    & \quad fluency              & 9.4 & 0.0 & 0.0 & 3.7 & 86.9 \\
               &    & Reflection, all criteria   & 23.2 & 6.3 & 21.3 & 11.8 & 58.7 \\
\addlinespace[1pt]
               & SF & Preference, overall        & 69.2 & 30.8 & 30.8 & 0.0 & 0.0 \\
               &    & Reflection: consistency    & 22.7 & 1.2 & 4.9 & 10.2 & 66.0 \\
               &    & \quad informativeness      & 68.6 & 7.3 & 9.7 & 18.9 & 5.1 \\
               &    & \quad coherence            & 14.5 & 0.2 & 1.1 & 10.3 & 75.0 \\
               &    & \quad fluency              & 11.2 & 0.0 & 0.0 & 3.8 & 85.0 \\
               &    & Reflection, all criteria   & 29.2 & 2.2 & 6.9 & 10.8 & 57.8 \\

\bottomrule
\end{tabular*}
\caption{Winner coverage and strict preference-reversal rates for the \emph{Equal}-enabled vanilla preference and reflection conditions, in percent of swap-pairs (valid $N=1{,}200$ per criterion and $4{,}800$ pooled, except DeepSeek-V3.1 reflection/SF: $N=1{,}199$ for informativeness, coherence and fluency, and $N=4{,}797$ pooled). \emph{Winner coverage} is the share with the same summary preferred in both orders, \emph{Rev.} the share with the same slot in both orders (the rate of Table~\ref{tab:strict_reversal_ablation}), \emph{Rev.$\mid$dec.}\ the biased share of the pairs decided A/B in both orders, \emph{Tie/preference disagreement} the share with an Equal verdict in exactly one order, and \emph{Stable abstention} the share with Equal verdicts in both orders. Winner coverage, strict reversals, Tie/preference disagreement and Stable abstention sum to $100\%$ up to rounding. AF = article-first, SF = summaries-first.}
\label{tab:winner_coverage}
\end{table*}

\subsection{Per-criterion bias without reflection}\label{app:per_criterion_norefl}
Table~\ref{tab:per_criterion_norefl} gives the no-reflection counterpart of Table~\ref{tab:eigenbench_per_criterion}. Informativeness still carries the highest bias rate in seven of the eight judge--placement cells -- the exception is GPT-4o-mini under summaries-first, where coherence ($57.9\%$) edges it out -- so the localization is not manufactured by the reflection step. The separation from the two form criteria is nevertheless much wider with reflection than without (for DeepSeek~V3.1/AF, $24.3\%$ against $0.2\%$ on coherence with reflection and $18.6\%$ against $10.9\%$ without; for GPT-OSS-120b/AF, $10.7\%$ against $0.2\%$ versus $15.5\%$ against $0.6\%$), which is the direction the account predicts, since what reflection adds is the judge's willingness to call fluency and coherence indistinguishable when they are. Where the bias lives is therefore a property of how discriminable the criterion is, and reflection sharpens that pattern rather than creating it.

\begin{table}[h]
\small
\setlength{\tabcolsep}{3pt}
\begin{tabular*}{\columnwidth}{@{\extracolsep{\fill}} llrrrr}
\toprule
Judge & Plc. & Cons. & Info. & Coh. & Flu. \\
\midrule
GPT-4o-mini       & AF &  3.5\% & 48.6\% & 31.0\% &  7.1\% \\
              & SF & 18.5\% & 51.1\% & 57.9\% & 28.5\% \\
GPT-5-mini        & AF &  4.6\% &  8.3\% &  1.5\% &  0.1\% \\
              & SF & 20.3\% & 31.6\% &  7.0\% &  1.1\% \\
GPT-OSS-120b  & AF &  0.3\% & 15.5\% &  0.6\% &  0.3\% \\
              & SF &  0.6\% & 20.3\% &  0.4\% &  0.3\% \\
DeepSeek-V3.1 & AF &  6.4\% & 18.6\% & 10.9\% &  1.3\% \\
              & SF & 18.7\% & 53.3\% & 11.4\% &  5.3\% \\
\bottomrule
\end{tabular*}
\caption{EigenBench per-criterion strict preference-reversal rates without the reflection step, with the \emph{Equal} option allowed (AF = article-first, SF = summaries-first).}
\label{tab:per_criterion_norefl}
\end{table}

\subsection{Summarizer-level preference scores}\label{app:absolute_scores}
\begin{figure}[h]
\begin{center}
\includegraphics[width=0.9\linewidth]{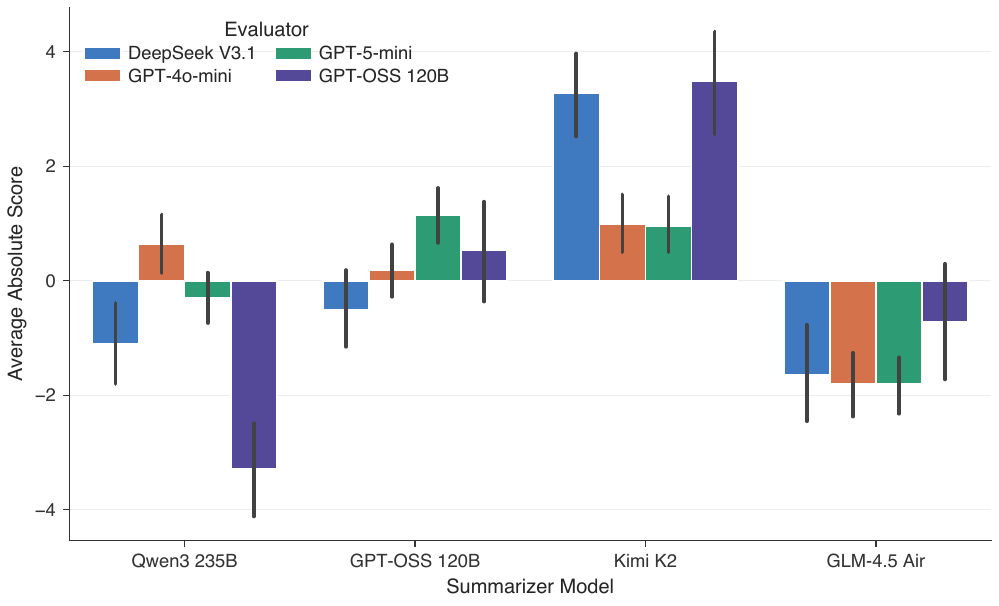}
\end{center}
\caption{Net preference score $S_J(M)$ per summarizer model $M$ (x-axis) and per judge $J$ (color), aggregated from Equal-enabled vanilla preference verdicts on the 200 articles (4 summarizer models, all 6 model pairs, summaries-first placement, signed $1$--$3$ preference strengths summed over opponents and articles). Higher = more often preferred. Error bars are bootstrap 95\% CIs across articles. Models are roughly ordered by mean rank. Judges largely agree on the top model but disagree about GPT-OSS-120b.}
\label{fig:absolute_comparison}
\end{figure}

Figure~\ref{fig:absolute_comparison} aggregates the Equal-enabled vanilla preference verdicts into a net preference score per summarizer and judge. We report it for completeness. Our focus is on the judges rather than the ranking of the four summarizers, and Appendix~\ref{app:judge_agreement} shows that the agreement underlying these rankings is only fair-to-moderate. We therefore do not draw model-selection conclusions from it.

\subsection{Inter-evaluator agreement on head-to-head verdicts}
\label{app:judge_agreement}
\begin{figure}[h]
\begin{center}
\includegraphics[width=0.85\linewidth]{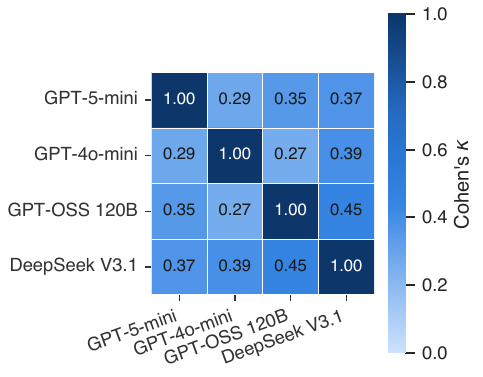}
\end{center}
\caption{Inter-evaluator agreement on head-to-head verdicts under the Equal-enabled vanilla preference protocol. Each cell is Cohen's $\kappa$ between two judges over the $1{,}200$ shared (article, model-pair) head-to-heads, after pooling the two summary orderings into a single net verdict in $\{A, \mathrm{Equal}, B\}$. Summaries-first placement, $200$ articles.}
\label{fig:judge_agreement_heatmap}
\end{figure}

Figure~\ref{fig:judge_agreement_heatmap} reports pairwise agreement between the four panel judges on individual head-to-head verdicts from the Equal-enabled vanilla preference protocol (Appendix~\ref{app:pairwise_prompt}). For each judge, each article, and each unordered pair of summarizer models, we pool the two orderings into a single net verdict in $\{A, \mathrm{Equal}, B\}$ by summing the signed preference-strength scores across both orderings, where $A$ and $B$ denote the two summarizers in the pair. Each cell shows Cohen's $\kappa$ between the two judges over the $1{,}200$ shared head-to-heads ($200$ articles $\times 6$ model pairs, with orderings pooled). Agreement is uniformly in the fair-to-moderate range ($\kappa \in [0.29, 0.52]$). Even with the \emph{Equal} option enabled and both orderings pooled, the four LLM judges agree on which summary wins a head-to-head only modestly more often than chance, despite all four producing broadly similar net rankings (Figure~\ref{fig:absolute_comparison}). In other words, judges concur on the aggregate ordering of summarizers far better than on the underlying per-judgment decisions that produce it.

\subsection{Licenses and intended use of artifacts}\label{app:licenses}
All artifacts used in this work are employed for non-commercial academic research, consistent with their original access conditions. The 200 source articles are drawn from The Guardian's Science pillar and are accessed through the publicly documented Guardian Open Platform API under its developer terms (free non-commercial tier). To respect The Guardian's content license, we do not redistribute the full article text or the LLM-generated summary texts. The released package contains article identifiers/URLs, judge verdicts, derived cosine and Smatch similarity caches, aggregated statistics, evaluation prompts, and code. The stored verdicts and caches support recomputation of the corresponding analyses without new model calls. Readers with appropriate API access can retrieve articles and generate new summaries. The open-weight summarizer models and their licenses are GPT-OSS-120b (Apache 2.0), Qwen3-235B-A22B-Instruct (Apache 2.0), GLM-4.5-Air (MIT), and Kimi-K2-Instruct (Modified MIT for research and commercial use). The judge models are accessed through hosted inference APIs (OpenAI for GPT-4o-mini and GPT-5-mini; DeepSeek for DeepSeek-V3.1; an open-weight provider for GPT-OSS-120b) and used within their published terms of service for evaluation research. Among the tooling, amrlib and its BART-large parser release~\cite{jascob2022amrlib}, as well as Smatch, are distributed under the MIT license. OpenAI \texttt{text-embedding-3-large} is used through the OpenAI API\@. The released code and verdicts are made available for research use only.

\paragraph{Personally identifying information and offensive content.} The corpus consists of professionally edited Guardian Science articles, a low-risk domain in which named individuals are public figures (researchers, officials, public commentators) mentioned in their professional capacity rather than as private individuals. The Guardian's editorial process already filters offensive content, and spot checks of the 200 articles and of the generated summaries did not surface any offensive or sensitive material. We do not redistribute the full article text or the generated summary texts (see above), so no additional anonymization step was required.

\subsection{Compute and inference budget}\label{app:compute}
No model training or fine-tuning was performed in this work, and all summarizer and judge models were used purely for inference through hosted APIs. Summarizer models (GPT-OSS-120b, Qwen3-235B-A22B-Instruct, GLM-4.5-Air, Kimi-K2-Instruct) were called through an open-weight hosted-inference provider. Judges were called through the OpenAI API (GPT-4o-mini, GPT-5-mini) and the same open-weight provider (GPT-OSS-120b, DeepSeek-V3.1). The dominant cost is the pairwise-evaluation grid. One judging pass is $6$ unordered model pairs $\times$ $200$ articles $\times$ $2$ orderings $=2{,}400$ ordered verdicts per judge and placement ($19{,}200$ across both), and the six protocols of Table~\ref{tab:master_bias} bring the total to $\sim\!128{,}000$ judge calls, including $12{,}800$ reflection generations. The stability check of Appendix~\ref{app:repeats} adds $16{,}200$ calls, and summary generation adds $\approx\!2{,}750$ calls (mean DVR trials per model in Table~\ref{tab:avg_iterations} $\times$ $200$ articles, first drafts included). AMR parsing with amrlib's BART-large parser~\cite{jascob2022amrlib} ($\sim\!800$ summary graphs $+\,200$ article graphs) was run on a single NVIDIA A100 (40~GB) in well under one GPU-hour. The end-to-end wall-clock time for the full pipeline is dominated by hosted-API latency rather than local compute.

\subsection{Software versions}\label{app:versions}
The released code includes \texttt{scripts/check\_reproducibility.py}, which checks the pinned list of 200 article identifiers and their 0-based indexing in result files, the rendered judge prompts for each placement and tie setting against pinned digests, and the word counter and the $10\%\pm10\%$ band. It also checks the configured model aliases and decoding settings, the recorded EigenBench metadata against each file's judge, placement, reflection and tie setting, and the absence of \emph{Equal} verdicts in forced-choice preference files. The original full-grid outputs did not record the endpoint-resolved model snapshots, so these cannot be verified retrospectively. Repeated passes record the endpoint-reported model identifiers, which the checker checks for consistency across passes. Checks that require the withheld texts are reported as skipped, never as passing.

We use amrlib's \texttt{parse\_xfm\_bart\_large} v0.1.0 AMR parser, based on \texttt{facebook/bart-large}, using amrlib version 0.8.0~\cite{jascob2022amrlib}. We compute the released Smatch cache using \texttt{smatch} 1.0.4 with $5$ hill-climbing restarts and a deterministic wrapper that fixes the random seed and suppresses Smatch's internal reseeding, with document-level scores obtained by accumulating \texttt{get\_amr\_match} triple counts across the sentence-aligned AMR graph pairs of a summary pair and micro-averaging them into a single F-score (rather than concatenating the sentence graphs into one block, which yields slightly different values). We access \texttt{text-embedding-3-large} through the \texttt{openai} Python SDK. We use \texttt{scipy} 1.13, \texttt{statsmodels} 0.14, \texttt{numpy} 2.x, and \texttt{pandas} 2.x for statistics and bootstrap CIs, and \texttt{matplotlib} 3.x and \texttt{seaborn} 0.13 for plots. The code runs under Python 3.11.

\end{document}

%% file: figures/framework.tex
% Framework overview figure (4 panels), drawn in TikZ.
% Requires in the preamble:
%   \usepackage{tikz}
%   \usetikzlibrary{arrows.meta, positioning, shapes.geometric, calc, fit, backgrounds, decorations.pathreplacing}
%   \usepackage{xcolor}
% Include with:  \resizebox{\linewidth}{!}{\input{figures/framework}}

\definecolor{cFair}{HTML}{CDE7D4}%
\definecolor{cAdj}{HTML}{F5CDC9}%
\definecolor{cDist}{HTML}{C9DCF3}%
\definecolor{cEq}{HTML}{FAE5C3}%
\definecolor{cMixed}{HTML}{DDD0EB}%
\definecolor{cStage}{HTML}{EDEFF2}%
\definecolor{cLine}{HTML}{4A4A4A}%
\definecolor{cPanel}{HTML}{F7F8FA}%
\definecolor{cAccent}{HTML}{9E2B25}%

% Font sizes are pinned so the figure renders identically at any base font size.
\begingroup%
\renewcommand{\tiny}{\fontsize{5.5}{6.3}\selectfont}%
\renewcommand{\scriptsize}{\fontsize{7}{8}\selectfont}%
\renewcommand{\normalsize}{\fontsize{9.5}{11}\selectfont}%
\begin{tikzpicture}[
  x=1cm, y=1cm,
  font=\tiny,
  every node/.style={inner sep=2pt},
  box/.style={draw=cLine, line width=0.4pt, rounded corners=1.6pt, align=center,
              fill=cStage, minimum height=6mm},
  io/.style={box, fill=white},
  slot/.style={draw=cLine, line width=0.3pt, rounded corners=1.2pt, align=left,
               fill=white, minimum height=3.4mm, inner xsep=3pt, anchor=north west,
               text width=42mm, font=\tiny\ttfamily},
  cell/.style={draw=cLine, line width=0.3pt, rounded corners=1.2pt, align=center,
               fill=white, minimum height=3.4mm, minimum width=6mm, font=\tiny},
  arr/.style={-{Latex[length=1.5mm,width=1.2mm]}, draw=cLine, line width=0.4pt},
  lbl/.style={font=\tiny\itshape, inner sep=1.5pt},
  dlt/.style={font=\tiny\itshape, inner sep=1.5pt},
  panel/.style={draw=cLine!50, line width=0.5pt, rounded corners=2.5pt, fill=cPanel},
  ptitle/.style={font=\scriptsize\bfseries, anchor=north west},
  note/.style={draw=cLine!60, line width=0.35pt, rounded corners=1.5pt, fill=white,
               align=left, anchor=north west},
]

%% ============================================================ panel frames
\draw[panel] (0,0) rectangle (10.15,-7.32);
\draw[panel] (10.35,0) rectangle (16.2,-7.32);
\draw[panel] (0,-7.52) rectangle (10.15,-11.50);
\draw[panel] (10.35,-7.52) rectangle (16.2,-11.50);

\node[ptitle] at (0.15,-0.08) {(a)\;\; Length-controlled candidates and the two prompt layouts};
\node[ptitle] at (10.5,-0.08) {(b)\;\; Swap-pair outcome taxonomy};
\node[ptitle] at (0.15,-7.60) {(c)\;\; Factorial ablation: total order disagreement};
\node[ptitle] at (10.50,-7.60) {(d)\;\; Where order disagreement remains};

%% ============================================================ (a) Stage 1: DVR
\node[font=\scriptsize\itshape, anchor=north west] at (0.28,-0.45)
  {Stage 1 --- candidate-side length control (Divide--Verify--Refine) on 200 Guardian Science articles};

\node[io]  (art)  at (1.10,-1.35) [text width=14mm] {Article $a$};
\node[box] (gen)  at (3.25,-1.35) [text width=21mm] {\textbf{Divide} $\to$ \textbf{Generate}\\[-1pt]{\tiny target $L=10\%$ of $a$}};
\node[box] (ver)  at (5.55,-1.35) [text width=17mm] {\textbf{Verify}\\[-1pt]{\tiny $|w(s)-L|\le\tau$?}};
\node[box] (ref)  at (3.25,-2.34) [text width=24mm] {\textbf{Refine}\\[-1pt]{\tiny numeric feedback $\Delta w$ only}};
\node[io]  (pool) at (8.35,-1.35) [text width=27mm]
  {Length-matched summaries\\[-1pt]{\tiny 4 summarizers $\times$ 200 articles}};

\draw[arr] (art) -- (gen);
\draw[arr] (gen) -- (ver);
\draw[arr] (ver) -- node[lbl,above,pos=0.5]{yes} (pool);
\draw[arr] (ver.south) |- node[lbl,pos=0.3,right]{no} (ref.east);
\draw[arr] (ref.north) -- node[lbl,right,pos=0.45]{revise} (gen.south);

\node[note, text width=30mm, text=cAccent] at (6.55,-1.81)
  {\textbf{Verbosity channel closes:} the longer summary wins
   $\mathbf{74\%\!\to\!57\%}$ of decided pairs.};

%% ------------------------------------------------ swap band
\node[draw=cLine, dashed, line width=0.35pt, rounded corners=1.5pt, fill=white,
      text width=88mm, align=center, anchor=north] (swap) at (4.90,-2.79)
  {every unordered pair $\{s_A,s_B\}$ is judged \textbf{twice} ---
   $(s_A,s_B)\!\to\!c_{\mathrm{fwd}}$ and $(s_B,s_A)\!\to\!c_{\mathrm{rev}}$
   --- under \textbf{both} layouts below};
\draw[arr] (pool.east) -- ++(0.25,0) |- (swap.east);

%% ------------------------------------------------ prompt cards
% ---- Article-first card
\draw[draw=cLine, line width=0.5pt, rounded corners=2pt, fill=white]
  (0.35,-3.51) rectangle (5.05,-6.69);
\node[font=\scriptsize\bfseries, anchor=north west] at (0.5,-3.66) {Article-first (AF)};

\node[slot, fill=cStage] at (0.5,-4.18)
  {[System] You are comparing two different summaries of the same text.};
\node[slot] at (0.5,-4.75)
  {Original Article: \{\{article\}\}};
\node[slot, fill=cAdj] at (0.5,-5.17)
  {Summary A : \{\{summary\_A\}\}};
\node[slot, fill=cDist] at (0.5,-5.59)
  {Summary B : \{\{summary\_B\}\}};
\node[slot, fill=cStage] at (0.5,-6.01)
  {Which do you prefer? If you like them equally, select "Equal".};

% ---- Summaries-first card
\draw[draw=cLine, line width=0.5pt, rounded corners=2pt, fill=white]
  (5.30,-3.51) rectangle (10.00,-6.69);
\node[font=\scriptsize\bfseries, anchor=north west] at (5.45,-3.66) {Summaries-first (SF)};

\node[slot, fill=cStage] at (5.45,-4.18)
  {[System] You are comparing two different summaries of the same text.};
\node[slot, fill=cDist] at (5.45,-4.75)
  {Summary A : \{\{summary\_A\}\}};
\node[slot, fill=cAdj] at (5.45,-5.17)
  {Summary B : \{\{summary\_B\}\}};
\node[slot] at (5.45,-5.59)
  {Original Article: \{\{article\}\}};
\node[slot, fill=cStage] at (5.45,-6.01)
  {Which do you prefer? If you like them equally, select "Equal".};

\node[anchor=north west, align=left, text=cAccent, text width=95mm] at (0.4,-6.84)
  {\textbf{Red = the slot adjacent to the article} --- fixed by the layout, not by the judge.};

%% ============================================================ (b) outcome taxonomy
\node[font=\scriptsize\bfseries] at (11.30,-0.72) {$c_{\mathrm{fwd}}$};
\node[font=\scriptsize\bfseries] at (12.15,-0.72) {$c_{\mathrm{rev}}$};

\newcommand{\pairrow}[4]{%
  \node[cell, fill=#4] at (11.30,#1) {#2};
  \node[cell, fill=#4] at (12.15,#1) {#3};}

% Both strict reversals and mixed Equal/preference outcomes are order-dependent.
\draw[draw=cLine!65, dashed, line width=0.4pt, rounded corners=2pt]
  (10.65,-1.84) rectangle (16.00,-4.38);
\node[anchor=north west, font=\tiny\bfseries, fill=cPanel, inner sep=2pt]
  at (10.82,-1.77) {Order disagreement};

\pairrow{-1.08}{A}{B}{cFair}
\pairrow{-1.50}{B}{A}{cFair}
\pairrow{-2.35}{A}{A}{cAdj}
\pairrow{-3.01}{B}{B}{cDist}
\pairrow{-3.66}{A/B}{E}{cMixed}
\pairrow{-4.10}{E}{A/B}{cMixed}
\pairrow{-4.82}{E}{E}{cEq}

\node[box, fill=cFair, text width=30mm, anchor=west] at (12.85,-1.29)
  {\textbf{Same model twice}\\[-1pt]{\tiny order-independent preference}};
\node[box, fill=cAdj, text width=30mm, anchor=west] at (12.85,-2.35)
  {\textbf{Slot \texttt{A} twice}\\[-1pt]{\tiny \emph{primacy} (article-adjacent under AF)}};
\node[box, fill=cDist, text width=30mm, anchor=west] at (12.85,-3.01)
  {\textbf{Slot \texttt{B} twice}\\[-1pt]{\tiny \emph{recency} (article-adjacent under SF)}};
\node[box, fill=cMixed, text width=30mm, anchor=west] at (12.85,-3.88)
  {\textbf{Tie/preference disagreement}\\[-1pt]{\tiny Equal in exactly one order}};
\node[box, fill=cEq, text width=30mm, anchor=west] at (12.85,-4.82)
  {\textbf{Stable abstention}\\[-1pt]{\tiny Equal in both orders}};

\draw[decorate, decoration={brace, amplitude=2.5pt}, draw=cLine, line width=0.3pt]
  (12.65,-0.90) -- (12.65,-1.68);
\draw[decorate, decoration={brace, amplitude=2.5pt}, draw=cLine, line width=0.3pt]
  (12.65,-3.48) -- (12.65,-4.28);
\draw[arr] (12.60,-2.35) -- (12.83,-2.35);
\draw[arr] (12.60,-3.01) -- (12.83,-3.01);
\draw[arr] (12.60,-4.82) -- (12.83,-4.82);

\node[note, text width=53mm] at (10.5,-5.30)
  {\textbf{Article-adjacency, not primacy or recency.} The article-adjacent
   slot carries the larger rate in \textbf{7 of 8} judge$\,\times\,$layout cells.};

\node[note, text width=53mm] at (10.5,-6.15)
  {\textbf{Position is a low-signal fallback.} Biased vs.\ consistent pairs:
   no similarity separation ($d\!=\!0.11$ cos., $0.04$ Smatch).
   Stable abstention vs.\ decided: pooled $d\!=\!0.38$ / $0.31$.};

%% ============================================================ (c) ablation
\node[font=\scriptsize\bfseries, anchor=south] at (3.00,-8.39) {forced A\,/\,B};
\node[font=\scriptsize\bfseries, anchor=south] at (7.20,-8.39) {$+$ Equal option};
\node[font=\scriptsize\bfseries, anchor=east, align=right] at (1.55,-8.92) {no\\reflection};
\node[font=\scriptsize\bfseries, anchor=east, align=right] at (1.55,-10.37) {$+$\,reflection\\{\tiny per criterion}};

\node[box, fill=cAdj, minimum width=24mm, minimum height=8mm, text width=22mm] (c00) at (3.00,-8.92)
  {\textbf{\normalsize 32.8--82.2\%}\\[1pt]{\tiny unmitigated}};
\node[box, fill=cEq, minimum width=24mm, minimum height=8mm, text width=22mm] (c01) at (7.20,-8.92)
  {\textbf{\normalsize 19.1--61.1\%}};
\node[box, fill=cAdj!50, minimum width=24mm, minimum height=8mm, text width=22mm] (c10) at (3.00,-10.37)
  {\textbf{\normalsize 26.7--63.2\%}};
\node[box, fill=cFair, minimum width=24mm, minimum height=8mm, text width=22mm] (c11) at (7.20,-10.37)
  {\textbf{\normalsize 11.4--36.4\%}\\[1pt]{\tiny both mitigations}};

\draw[arr, line width=0.6pt] (c00) -- node[dlt, above]{$-29.1$ to $+10.4$} (c01);
\draw[arr, line width=0.6pt] (c10) -- node[dlt, above]{$-29.5$ to $-6.8$} (c11);
\draw[arr, line width=0.6pt] (c00) -- node[dlt, left]{$-19.0$ to $+8.1$} (c10);
\draw[arr, line width=0.6pt] (c01) -- node[dlt, right]{$-35.0$ to $-7.8$} (c11);

\node[anchor=north, font=\tiny, align=center, text width=88mm] at (5.00,-10.85)
  {\textcolor{cAccent}{\textbf{Both together yield the lowest disagreement in 8 of 8 settings; adding either to the other lowers disagreement in all 8.}}};

% ---- (d) per-criterion bars
\node[font=\tiny\itshape, anchor=north west, text width=55mm] at (10.50,-7.98)
  {total order disagreement per criterion; median over the 8 judge\,$\times$\,layout cells, both mitigations on};

\begin{scope}[shift={(12.72,-8.67)}]
  \foreach \n/\v/\y in {Informativeness/42.7/0, Consistency/15.3/-0.45, Coherence/10.3/-0.90, Fluency/4.0/-1.35}{
    \node[anchor=east, font=\tiny] at (-0.08,\y) {\n};
    \fill[cLine!25, draw=cLine, line width=0.3pt] (0,\y-0.105) rectangle ({0.02+\v*0.055},\y+0.105);
    \node[anchor=west, font=\tiny] at ({0.14+\v*0.055},\y) {\v\%};
  }
  \draw[draw=cLine!60, line width=0.3pt] (0,0.25) -- (0,-1.6);
\end{scope}

\node[anchor=north west, font=\tiny, text width=55mm, align=left] at (10.50,-10.42)
  {Disagreement is highest for informativeness and remains nonzero across all four criteria.
   The appendix separates strict reversals from tie/preference disagreement.};

\end{tikzpicture}
\endgroup%

%% file: figures/reflection.tex
% Reflection-based (EigenBench-style) judging protocol -- appendix figure.
% Requires the same preamble as figures/framework.tex.
% Include with:  \resizebox{\columnwidth}{!}{\input{figures/reflection}}
%
\definecolor{cFair}{HTML}{CDE7D4}%
\definecolor{cAdj}{HTML}{F5CDC9}%
\definecolor{cEq}{HTML}{FAE5C3}%
\definecolor{cStage}{HTML}{EDEFF2}%
\definecolor{cLine}{HTML}{4A4A4A}%
%
% Font sizes are pinned so the figure renders identically at any base font size.
\begingroup%
\renewcommand{\tiny}{\fontsize{5.5}{6.3}\selectfont}%
\renewcommand{\scriptsize}{\fontsize{7}{8}\selectfont}%
\begin{tikzpicture}[
  x=1cm, y=1cm,
  font=\tiny,
  every node/.style={inner sep=2pt},
  box/.style={draw=cLine, line width=0.4pt, rounded corners=1.6pt, align=center,
              fill=cStage, minimum height=6mm},
  io/.style={box, fill=white},
  res/.style={box, minimum height=7mm},
  arr/.style={-{Latex[length=1.5mm,width=1.2mm]}, draw=cLine, line width=0.4pt},
  lbl/.style={font=\tiny\itshape, inner sep=1.5pt},
]

\node[io]  (pair) at (1.90,-0.42) [text width=32mm]
  {Length-matched pair $(s_A,s_B)$\\[-1pt]{\tiny from Stage~1}};
\node[io]  (con)  at (5.75,-0.42) [text width=34mm]
  {Constitution $C$\\[-1pt]{\tiny informativeness, consistency, coherence, fluency}};

\node[box] (refl) at (3.85,-1.75) [text width=60mm]
  {\textbf{Reflect} --- per criterion, each summary alone\\[-1pt]
   {\tiny $r_A = J(a,s_A,C)$ \qquad $r_B = J(a,s_B,C)$ \\ no winner is named, so $r_A$ and $r_B$ are order-independent}};

\node[box] (fwd) at (1.95,-3.30) [text width=34mm]
  {\textbf{Compare} forward\\[-1pt]{\tiny $(s_A,r_A)$ vs.\ $(s_B,r_B)\ \to\ c_{\mathrm{fwd}}$}};
\node[box] (rev) at (5.75,-3.30) [text width=34mm]
  {\textbf{Compare} reverse\\[-1pt]{\tiny $(s_B,r_B)$ vs.\ $(s_A,r_A)\ \to\ c_{\mathrm{rev}}$}};

\node[box] (swap) at (3.85,-4.55) [text width=30mm]
  {\textbf{Swap check} {\tiny(per criterion)}};

\node[res, fill=cFair, text width=20mm] (verd) at (1.30,-6.15)
  {\textbf{Verdict $A$\,/\,$B$}\\[-1pt]{\tiny order-invariant}};
\node[res, fill=cEq,   text width=20mm] (tie)  at (3.85,-6.15)
  {\textbf{\emph{Equal}}\\[-1pt]{\tiny judge abstains}};
\node[res, fill=cAdj,  text width=20mm] (bias) at (6.40,-6.15)
  {\textbf{Positional bias}\\[-1pt]{\tiny same slot wins twice}};

\draw[arr] (pair.south) -- ++(0,-0.30) -| (refl.north -| pair.south);
\draw[arr] (con.south)  -- ++(0,-0.30) -| (refl.north -| con.south);
\draw[arr] (refl.south -| fwd.north) -- (fwd.north);
\draw[arr] (refl.south -| rev.north) -- (rev.north);
\draw[arr] (fwd.south) |- ++(0,-0.28) -| (swap.north);
\draw[arr] (rev.south) |- ++(0,-0.28) -| (swap.north);
\draw[draw=cLine, line width=0.4pt] (swap.south) -- (3.85,-5.45);
\draw[draw=cLine, line width=0.4pt] (1.30,-5.45) -- (6.40,-5.45);
\draw[arr] (1.30,-5.45) -- node[lbl, right]{same model}      (verd.north);
\draw[arr] (3.85,-5.45) -- node[lbl, right]{tie either order} (tie.north);
\draw[arr] (6.40,-5.45) -- node[lbl, left]{same slot}         (bias.north);

\end{tikzpicture}
\endgroup%